\documentclass{article}

\PassOptionsToPackage{numbers, compress}{natbib}

\usepackage[main, final]{neurips_2026}

\usepackage[utf8]{inputenc} 
\usepackage[T1]{fontenc}    
\usepackage{url}            
\usepackage{booktabs}       
\usepackage{amsfonts}       
\usepackage{nicefrac}       
\usepackage{microtype}      
\usepackage{xcolor}         

\usepackage[utf8]{inputenc} 
\usepackage[T1]{fontenc}    
\usepackage[breaklinks=true,  colorlinks,  bookmarks=false]{hyperref}  
\usepackage{hyperref}       
\usepackage{url}            
\usepackage{booktabs}       
\usepackage{amsfonts}       
\usepackage{nicefrac}       
\usepackage{microtype}      
\usepackage{xcolor}         
\usepackage{array}
\usepackage{makecell}

\usepackage{graphicx}
\usepackage{amsmath}
\usepackage{multirow}
\usepackage{caption}
\usepackage{subcaption}
\usepackage{bm}
\usepackage{soul}

\usepackage{enumitem}
\usepackage{minitoc}
\usepackage{fontawesome} 
\usepackage{marvosym}

\usepackage{wrapfig}

\usepackage{amsmath}
\usepackage{amssymb}
\usepackage{amsthm}
\theoremstyle{plain}

\usepackage{multirow} 
\usepackage{multicol} 
\usepackage{extarrows}
\usepackage{float} 
\usepackage{makecell}

\usepackage{algorithm}
\usepackage[noend]{algpseudocode}

\usepackage{wrapfig}
\usepackage{graphicx}

\usepackage{comment}
\usepackage{array}
\usepackage[table]{xcolor}
\newtheorem{theorem}{Theorem}

\usepackage{colortbl}

\definecolor{oursblue}{RGB}{240,240,240}

\definecolor{mydarkblue}{rgb}{0,0.08,0.45}
\definecolor{mydarkgreen}{RGB}{0, 139, 69}

\title{Chatter Attack: Resource Consumption Attack for Large Language Models}

\renewcommand\footnotemark{}
\author{
Yibo Miao$^{1,2\star}$, Yichuan Cao$^{2\star}$, Xiao-Shan Gao$^{2\dagger}$, Yinpeng Dong$^{1,3\dagger}$ 
\thanks{$^\star$Equal contribution. $^{\dagger}$ Corresponding author.
\;\;$^{\ddagger}$ miaoyibo@amss.ac.cn, dongyinpeng@tsinghua.edu.cn}\\
  $^{1}$T-STAR Lab, College of AI, Tsinghua University, Beijing 100084, China\\
  $^{2}$State Key Laboratory of Math. Sci., Academy of Mathematics and Systems Science,\\ Chinese Academy of Sciences, Beijing 100190, China \;\; $^{3}$Shanghai Qi Zhi Institute
}

\begin{document}

\maketitle

\vspace{-2ex}
\begin{abstract}

LLMs are powerful but incur substantial inference cost, which can be exploited by malicious users to induce overly long outputs, increasing latency and operational expense. 
Prior resource consumption attacks either coerce repetition or suppress the end-of-sequence token by lowering its logit value.
However, with autoregressive, sampling-based generation, the search space grows exponentially with length and small deviations or unstable trajectories can derail optimization, limiting attack effectiveness and stability.
To address this issue, we introduce \textbf{Chatter Attack}, a novel class of resource consumption attacks that optimize adversarial prompts to induce predefined short target phrases, thereby avoiding search-space explosion while still eliciting long responses. Specifically, Chatter Attack triggers self-doubt using short doubt-inducing phrases as the target, and sustains this state throughout generation via inference attention loss and global entropy loss.
To extend to black-box settings, we further propose a novel \textbf{Bayesian Prompt Optimization (BPO)} method, 
which models the objective function globally by constructing a discrete token kernel, efficiently exploring the entire search space using prior information.
Extensive experiments across multiple models and datasets show that our method outperforms baselines, with maximum response length increases to 31.5$\times$, and revealing practical resource-depletion risks 
for real-world LLM inference services.
  
\end{abstract}

\section{Introduction}

Large language models (LLMs) demonstrate strong performance across a wide range of real-world applications~\citep{zhao2023survey, chang2024survey, naveed2025comprehensive}. With chain-of-thought prompting~\citep{wei2022chain}, these models generate multi-step reasoning traces that decompose complex problems into structured step-by-step solutions~\citep{feng2023towards, wu2025more}. This has become a core feature of modern reasoning LLMs (e.g., OpenAI o1/o3~\citep{jaech2024openai}, DeepSeek-R1~\citep{guo2025deepseek}, QWQ~\citep{team2025qwq}), improving performance in domains such as mathematics~\citep{ahn2024large}, programming~\citep{yang2023coupling}, and logical reasoning tasks requiring careful intermediate inference and systematic verification~\citep{liu2025logical}.

These modern LLMs typically contain billions of parameters, and each inference generation can require substantial computational cost. Both NVIDIA and Amazon Web Services~\citep{patterson2021carbon, patterson2022carbon} indicate that the inference phase during deployment accounts for more than 90\% of the resource demands. 
The costs associated with supporting LLM inference services often surpass those of training~\citep{patel2024splitwise, li2024llm}, due to the growing and widespread user base. 
This creates a new security threat~\citep{feng2024llmeffichecker}: attackers can craft malicious inputs to force the model to produce unusually lengthy outputs, consuming excessive computational resources and increasing generation latency and operational costs. For applications offering free access (e.g., Deepseek-R1~\citep{guo2025deepseek}) or pay-per-use/subscription-based API access, such attacks pose a significant operational cost threat and can severely impact the availability for legitimate users. More critically, these attacks can lead to resource depletion and service disruption, thereby compromising the reliability, availability, and scalability of real-world LLM deployments~\citep{gao2024denial, geiping2024coercing}.

Resource consumption attacks were first studied by Sponge Samples~\citep{shumailov2021sponge} through maximizing activation norms to increase inference cost. NICGSlowDown \citep{chen2022nicgslowdown} manipulates token logits to slow decoder-based generation, but neither directly transfers to autoregressive LLMs.
\citet{geiping2024coercing} show that adversarial prompts can coerce repetition (e.g., 24$\times$ ``Hello There''), extending output length and latency. Likewise, Engorgio-style methods~\citep{dong2025engorgio, si2025excessive, yan2025bithydra, gao2024inducing, feng2024llmeffichecker} suppress $\langle\text{EOS}\rangle$ by minimizing its logit relative to other tokens, prolonging generations.
However, autoregressive sampling makes these objectives hard to optimize: the search space grows exponentially with sequence length, and a single anomalous token can derail the generation trajectory, degrading attack effectiveness and stability.

To address this issue, our core insight lies in targeting short sequences as the optimization objective to prevent exponential search space expansion during optimization, while still influencing the length of longer sequences. We introduce a novel attack called \textbf{Chatter Attack}.
Specifically, we introduce short phrases with doubtful semantics as the attack target at the beginning of the output sequence, aiming to induce a state of self-doubt in the model from the start. This encourages the model to continually reflect and generate more tokens. 
However, although the model falls into self-doubt at the start of the response, it doesn't necessarily affect token outputs after thousands of steps.
To mitigate this, we focus on the attention values of key reasoning tokens (e.g., "wait" and "alternative") 
and the subsequent tokens in the sequence, aiming to increase their attention towards the adversarially injected prefix tokens. By targeting short phrases with doubtful semantics, we ensure that the model maintains self-doubt throughout the entire sequence.
Furthermore, we enhance uncertainty by maximizing the entropy of the predicted token distribution 
at each generation step. 
This encourages the model to deviate from the expected output in its original normal state, facilitating deeper self-doubt.

In the white-box scenario,
we optimize adversarial prompt using Greedy Coordinate Gradient (GCG)~\citep{zou2023universal}.
However, GCG assumes the availability of gradient information, making it unsuitable for black-box scenarios. To overcome this limitation, 
we propose \textbf{Bayesian Prompt Optimization (BPO)}, a novel Bayesian optimization-based adversarial prompt optimization method, 
designed to effectively execute resource consumption attacks in black-box settings.
The core idea behind BPO is to treat the function values returned by each iteration of the optimization process as prior information, guiding the strategy for the next iteration. By employing Bayesian optimization~\citep{frazier2018tutorial}, BPO models the objective function globally, enabling efficient exploration of the entire search space. We construct a special kernel function tailored for discrete token optimization spaces, thus the BPO method 
can more effectively and efficiently minimize the objective function and identify adversarial prompts.

To validate the superiority of our Chatter Attack and Chatter-BPO Attack, we conduct extensive experiments across five datasets, including mathematical reasoning \citep{cobbe2021training, amini2019mathqa, lightman2023let}, coding \citep{chen2021evaluating}, and question-answering \citep{wang2024mmlu}, on multiple modern LLMs, comparing with over ten baselines. The results show that both attacks significantly outperform all baselines. Notably, Chatter Attack achieves an impressive maximum response length increase of up to 31.5$\times$ on the GSM-8K dataset, while Chatter-BPO Attack reaches a maximum increase of 18.5$\times$.
Additionally, a local LLM service is deployed on NVIDIA A800 GPU, and our attacks are executed under controlled conditions, simulating real-world deployment and attack scenarios.
Our methods increase response latency and degrade service quality, with extreme cases leading to GPU exhaustion and complete service failure, demonstrating their practicality.
Further experiments with various mitigation strategies and ablation studies confirm the robustness of Chatter Attack and Chatter-BPO Attack. 
Our results highlight the resource depletion risks of large models and online LLM services, underscoring the need for defensive mechanisms to detect and mitigate excessive inference caused by adversarial prompts in real-world scenarios.

\section{Threat Model} 

\textbf{Attack Scenario.}
Our attack aims to induce large language models to output as many tokens as possible, thereby increasing inference costs. 
For LLM API service providers, malicious users or business competitors may intentionally trigger resource consumption attacks to induce the generation of excessively long outputs. This results in a significant computational load on the servers, leading to higher operational costs and excessive response delays that impact the availability of services for legitimate users, resembling a denial-of-service (DoS) attack.
For legitimate users, the query might involve polluted external data sources (e.g., documents, code repositories on GitHub), or it could be manipulated by untrustworthy third-party API proxies, intentionally triggering resource consumption attacks. These attacks lead to additional inference expenses that significantly exceed expectations. 
Furthermore, in scenarios involving large models interacting with the physical world, excessive latency can prevent embodied robots or autonomous vehicles from responding in time, creating safety risks for users and others in the vicinity.
Finally, the increased complexity of outputs may lead to unnecessary carbon emissions.

\textbf{Attacker goal.}
An LLM maps an input sequence $x$ to an output sequence $y = \{y_1, y_2, \dots, y_N\}$, where $y_i$ denotes the $i$-th token and $N$ represents the length of the output sequence. The attack objective is to append an adversarial prefix $a$ to the original input $x$, in order to maximize the output length $N$.
The LLM generates tokens in an autoregressive manner. After the Softmax layer, the probability distribution for the $i$-th generated token is represented as $f(a; x; y_1, y_2, \dots, y_{i-1})$, with the probability of generating $y_i$ as $p(y_i~|~a; x; y_1, y_2, \dots, y_{i-1})$. 
The probability of halting at the $i$-th token is $p(\langle\text{EOS}\rangle \mid a; x; y_1, y_2, \dots, y_{i-1})$, where $\langle\text{EOS}\rangle$ denotes the end-of-sequence token.

\textbf{Attacker capability.}
We consider both white-box and black-box scenarios. In the white-box scenario, the attacker has full access to the target model's architecture, parameters, and gradients.
This setting is supported by open-source models deployed in user-facing services \citep{llama_companies,guo2025deepseek,agarwal2025gpt}. 
In the black-box scenario, the attacker has black-box access to the target model and a restricted number of queries.

\section{Chatter Attack}

To increase the output length of LLMs, previous attack methods~\citep{geiping2024coercing, dong2025engorgio, si2025excessive, yan2025bithydra, gao2024inducing, feng2024llmeffichecker} either forced the model to repeat specific content (e.g., setting the attack output target as 24 repetitions of "Hello There") or minimized the logits of the end-of-sequence token relative to each output token, thereby suppressing the appearance of the $\langle\text{EOS}\rangle$ token.
However, these approaches face challenges due to the exponential growth of the search space with the increase in the number of tokens, which makes it difficult to effectively optimize. Specifically, as LLM generation is autoregressive and 
sampling-based, 
different random output sequences are generated for the same input sample. This complicates the optimization objective based on specific long sequences (i.e., 24 repetitions of "Hello There" or each non-$\langle\text{EOS}\rangle$ output token). The appearance of an anomalous token can 
cause generation to deviate from the target optimization trajectory, thus affecting the attack's effectiveness and stability.

To address this issue, our core insight lies in targeting short sequences to prevent the exponential expansion of the search space during optimization while still influencing the length of longer sequences. Specifically, we propose using short phrases with doubtful semantics as the attack target at the starting point of the output sequence, aiming to induce a state of self-doubt in the model from the outset, prompting it to continuously reflect and generate more tokens. 
Additionally, we optimize the attention values of key reasoning tokens (e.g., "wait" and "alternative") and the subsequent $n$ tokens in relation to the short doubt-inducing phrases. This enables the model to maintain a state of self-doubt throughout the entire sequence.

We propose \textbf{Chatter Attack}, which introduces three losses to optimize the prefix $a$: 
(1) Initial Doubt Loss, designed to induce self-doubt in the model at the beginning; 
(2) Inference Attention Loss, which encourages the model to maintain self-doubt throughout the entire sequence; and 
(3) Global Entropy Loss, which enhances output uncertainty to break the model's reliance on the original short output sequence. 

\subsection{Initial Doubt Loss}

\begin{wraptable}{r}{0.48\textwidth}
\centering\small
\setlength{\tabcolsep}{2pt}
\vspace{-3ex}
\caption{
We calculate the Perplexity (PPL) of several potential optimization target phrases on R1-7b, and the skeptical phrases exhibited lower PPL compared to other options, indicating that skeptical phrases better align with the model's expectations.
}
\label{tab:ppl}
\resizebox{\linewidth}{!}{
\begin{tabular}{cc}
\hline
Potential Optimization Target Phrases                      & PPL            \\ \hline
Okay, so I should think again and again                    & 229.20         \\
Okay, so I decide to ignore the question                   & 103.87         \\
Okay, so I think cats sleep most of their lives            & 330.63         \\
Okay, so I think I should leave 25\% of investments        & 154.25         \\
Okay, so I believe this is a logical paradox               & 141.92         \\
Okay, so I've found a new perspective on this              & 184.53         \\
Okay, so I should first guess the answer is 175            & 98.33          \\
Okay, so I need to add a twist                             & \textbf{81.97} \\ \hline
\vspace{-5ex}
\end{tabular}
}
\end{wraptable}

~\citet{zou2023universal} employed positive affirmations, such as “Sure, here is...” to bypass the safety mechanisms of LLMs and provoke unsafe behavior. 
Based on this, we aim to construct an initial phrase that meets the following objectives: 
(1) a longer subsequent output sequence. If the phrase places the language model in a state that encourages additional reasoning branches, it is likely to result in more output tokens. 
(2) ease of optimization. If the phrase follows the natural linguistic patterns of the model, the resulting outputs tend to lie in high-probability regions of the model’s output distribution, making it easier and smoother to optimize as the attack target. 
However, these two goals are in competition. Phrases that strongly induce longer outputs often push the model into less typical generation regimes, causing the output distribution to deviate from its original natural patterns. Such distributional shifts increase the variance and irregularity of the optimization objective, making the attack more difficult to optimize~\citep{jia2024improved}. Balancing this trade-off is therefore a key challenge.

We propose using short phrases with skeptical semantics (e.g., “Okay, I need to add a twist”) as the optimization target for the beginning of the output sequence. On one hand, this places the model in a state of self-doubt right from the start, prompting it to continuously reflect and output additional reasoning tokens. On the other hand, skepticism and reflection are common steps in normal outputs, aligning with the model’s original output patterns. To further validate this, we calculate the Perplexity (PPL) of several potential target phrases, with lower PPL indicating that the string better fits the model's expectations. As shown in Table~\ref{tab:ppl}, the skeptical phrases exhibit lower PPL compared to other options, making them suitable for optimization targets.

Thus, we select short skeptical phrases, denoted as $y^{*}_{1:M}$, as the optimization target, where $M$ is the token count of the phrase. We define the Initial Doubt Loss using the standard cross-entropy loss function:
\begin{equation}
\mathcal{L}_{\text{ID}} = -\sum_{m=1}^{M} \log p\!\left(y^{*}_m \mid a; x; y^{*}_{<m}\right)
\end{equation}
This loss function maximizes the probability of generating the specific target sequence $y^{*}_{1:M}$, effectively inducing the model into a state of self-doubt.

\subsection{Inference Attention Loss}
\label{sec:attention_loss}

\begin{wrapfigure}{r}{0.48\textwidth}
    \vspace{-3ex}
    \centering
    \includegraphics[width=\linewidth]{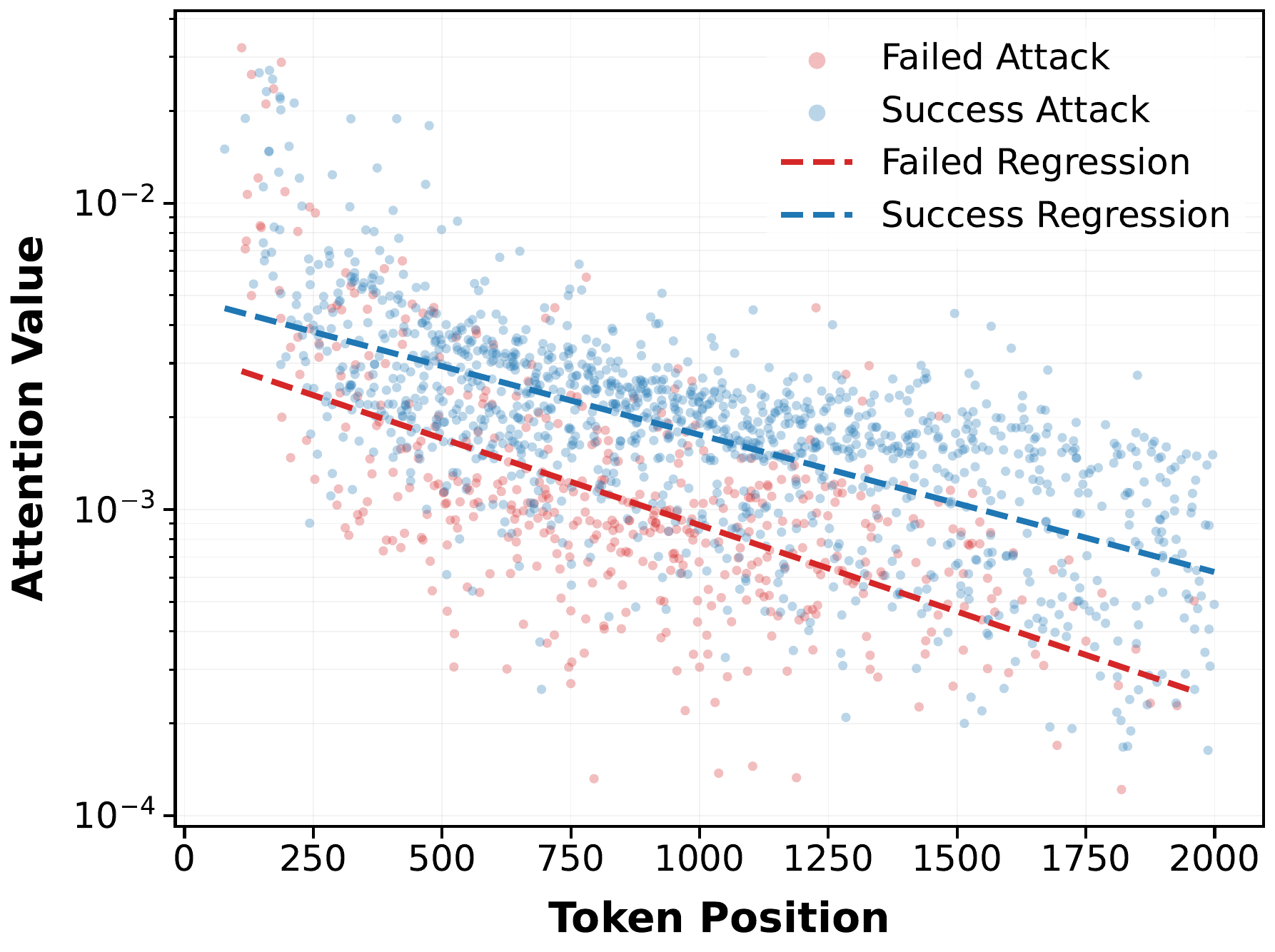}
    \caption{
    Visualization of how tokens, following key reasoning tokens at different positions on the GSM-8K dataset using R1-7b, assign attention weights toward the adversarially injected prefix tokens.
    Blue points indicate successful attack samples, while red are failed ones. The lines are regression lines.
    In failed attacks, attention values are lower, as the model fails to focus on the adversarial prefix tokens at key reasoning token positions during later output sequence positions.}
    \label{fig2}
\end{wrapfigure}

Although the model falls into self-doubt at the start of the response, it does not necessarily influence the token outputs after thousands of steps. 
We analyze how tokens following different key reasoning tokens at various positions exhibit attention weights toward the adversarially injected prefix tokens.
In fact, as shown in Figure~\ref{fig2}, when only the initial doubt loss is used, in cases where the attack fails, 
the tokens that follow key reasoning tokens (e.g., “wait” and “alternative”) in the later part of the sequence exhibit lower attention weights toward the adversarially injected prefix tokens $y^{*}_{1:M}$,
while in successful attack cases, the attention weights are higher.
Building on this insight, we aim to increase the attention scores of tokens following critical reasoning tokens, directing more attention towards the adversarially injected prefix tokens, 
ensuring that the model maintains awareness of the initial skeptical phrase throughout the entire output.

Specifically, we compute the attention map $A_{l,i}$ at the $l$-th layer for the $i$-th token w.r.t. the previous $i-1$ tokens:
\begin{equation}
A_{h,l,i} = \mathrm{softmax}\!\left(\frac{Q_{l,i}K_{l,i}^{\top}}{\sqrt{d_k}}\right)_{h}, 
A_{l,i}=\frac{1}{H}\sum_{h=1}^{H}A_{h,l,i},
\end{equation}
where $Q_{l,i}\in\mathbb{R}^{1\times d_k}$ is the query vector of the $i$-th token, $K_{l,i}\in\mathbb{R}^{(i-1)\times d_k}$ is the key matrix of the preceding tokens, $H$ is the number of heads, and $A_{h,l,i}$ represents the attention weight from the $h$-th attention head at layer $l$ for the $i$-th output token. Thus, $A_{l,i}^{(j)}$ denotes the attention weight assigned to the $j$-th token when generating the $i$-th token.
We define the Inference Attention Loss as:
\begin{equation}
\mathcal{L}_{\text{IA}} = -\sum_{l} \sum_{i \in \mathcal{U}} \sum_{j \in \mathcal{V}} A_{l,i}^{(j)},
\end{equation}
where $\mathcal{V} = \{idx~|~\text{token}_{idx} \in y^{*}_{1:M} \}$ is the set of indices for tokens from the initial skeptical phrase, and $\mathcal{U} = \{idx, idx+1, \dots, idx+n \mid \text{token}_{idx} \in \mathcal{T} \}$ is the set of indices for the critical reasoning tokens and the subsequent $n$ tokens, where $n$ defines the length of the optimizable short sequence and $\mathcal{T}$ is the set of critical reasoning tokens (e.g., "wait" and "alternative") as detailed in Appendix~\ref{App:experiment_settings}.
This loss function forces the selected token sequence to assign high attention to the initial skeptical phrase, thereby ensuring the model retains self-doubt throughout the entire inference process.

\subsection{Global Entropy Loss}
\vspace{-0.5ex}

\begin{wrapfigure}{r}{0.48\textwidth}
    \centering
    \vspace{-3.5ex}
    \includegraphics[width=\linewidth]{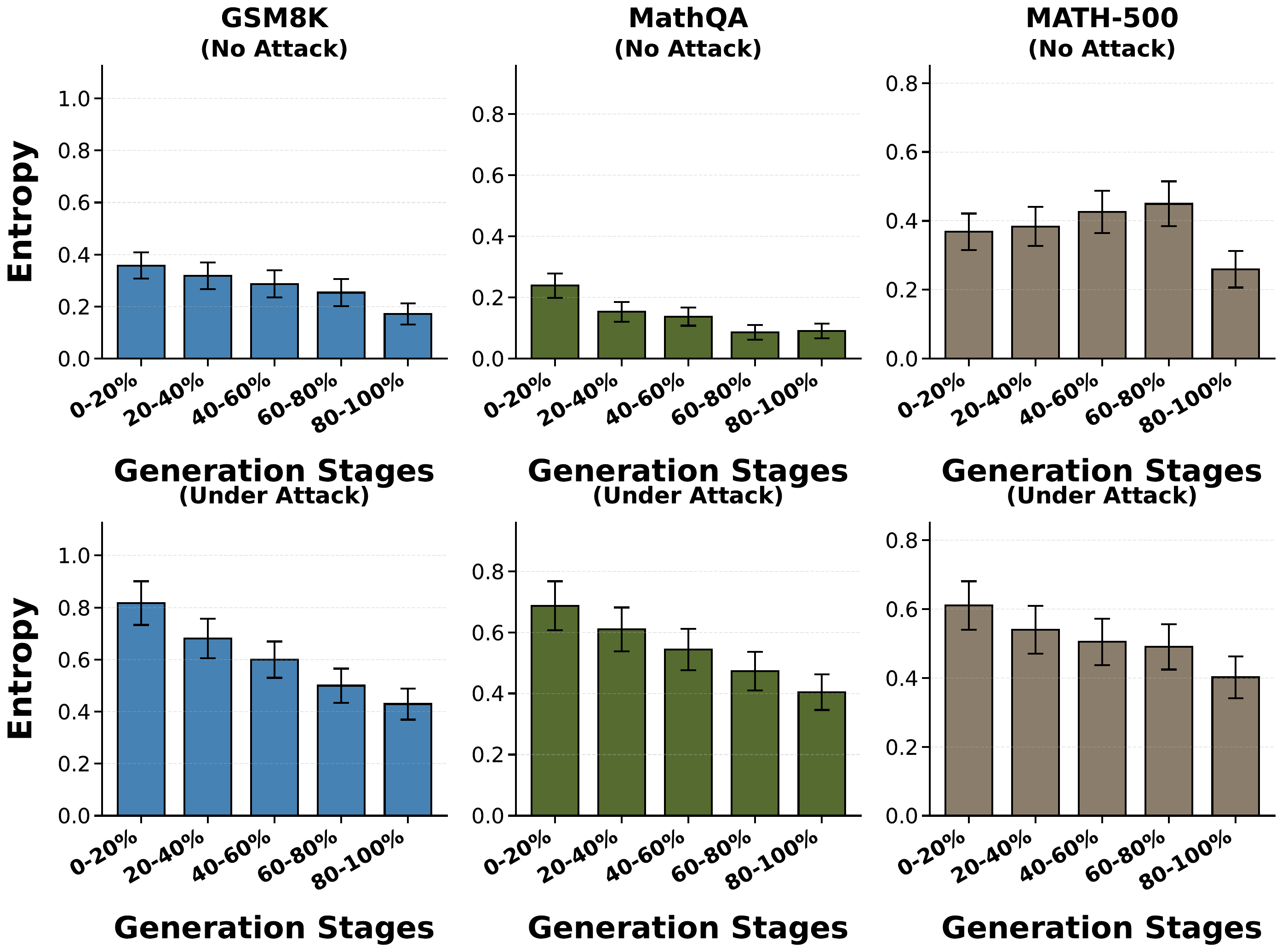}
    \caption{Visualization of uncertainty trajectories at different reasoning stages across diverse datasets on R1-7b, under benign and standard-attack settings. The x-axis is the \% of reasoning progress and the y-axis is the uncertainty. In later stages, the model focuses on the final answer and uncertainty decreases.
    Standard attacks increase uncertainty even without global entropy loss during autoregressive inference.
    }
    \vspace{-6ex}
    \label{fig3}
\end{wrapfigure}

Finally, we immerse the model deeper in the state of self-doubt, rather than allowing it to recover to a normal state. We compute uncertainty at different stages of reasoning across diverse datasets, under both benign and attacked settings. As shown in Figure~\ref{fig3}, in the later stages of inference, especially when approaching the final answer, the model's reasoning narrows around the correct answer, with its confidence increasing and uncertainty decreasing.
To counteract this, we propose enhancing the output uncertainty of generated tokens, steering the model's output away from the expected response in a normal state, facilitating a deeper induction into self-doubt. 
We achieve this by minimize the negative entropy of the predicted token:
\begin{equation}
\mathcal{L}_{\text{GE}} = \sum_{k=1}^{K} \sum_{i=1}^{|V|} f_i(a; x; y_{<k}) \log f_i(a; x; y_{<k}),
\end{equation}
where $f(a; x; y_{<k})$ represents the probability distribution of the $k$-th generated token, and $f_i(a; x; y_{<k})$ is its $i$-th component.
This loss function introduces a more uniform distribution and greater uncertainty for each token's prediction, allowing the model to break its dependency on the original output sequence and be induced into a state of self-doubt.


\subsection{Overall Objective Function}
\vspace{-0.5ex}

We combine the three loss functions described above into an overall objective function to minimize:
\begin{equation}
\mathcal{L} = \alpha \mathcal{L}_{\text{ID}} + \beta \mathcal{L}_{\text{IA}} + \lambda \mathcal{L}_{\text{GE}}.
\end{equation}
where $\alpha$, $\beta$, and $\lambda$ are the balancing hyperparameters.

\subsection{Theoretical Analysis}\label{sec:3-5}
\vspace{-0.5ex}

To explain how the three losses encourage longer outputs, we establish a theoretical connection between their weighted combination $\mathcal{L}_{\mathrm{proxy}}$ and a lower bound on the expected generation length $\mathbb{E}[N_B]$, motivating its use as a proxy for the expected stopping time.
We have the following theorem with a detailed proof provided in Appendix~\ref{app:theorem}:

\begin{theorem}[Surrogate bound on expected stopping time]
\label{thm:main}
Assume generation and loss evaluation use the same sampling distribution.
After the initial target phrase is generated, suppose every reachable
unfinished history has $\langle\text{EOS}\rangle$ probability at most $e^{-\gamma_{\mathrm c}}$, where
$\gamma_{\mathrm c}=-\widetilde\beta\mathcal{L}_{\text{IA}}
-\widetilde\lambda\mathcal{L}_{\text{GE}}-b-\varepsilon
\in[0,\log(B-M)]$. Then
\[
 \log\mathbb{E}[N_B]\geq-\mathcal L_{\mathrm{proxy}}-C,
 \qquad
 \mathcal L_{\mathrm{proxy}}
 =\mathcal{L}_{\text{ID}}
 +\widetilde\beta\mathcal{L}_{\text{IA}}
 +\widetilde\lambda\mathcal{L}_{\text{GE}}.
\]
Here $N_B=\min\{\tau,B\}$ counts decoding steps through the first EOS or
budget $B$; $M<B$ is the length of the $\langle\text{EOS}\rangle$-free output target
(denoted by $y^*_{1:M}$ in the main text);
$\widetilde\beta=\kappa_\Delta/(L|\mathcal U|)$ and
$\widetilde\lambda=1/K$, where $L$ counts layers, $\mathcal U$ is the
main-text set of selected attention positions, and $K$ counts entropy
positions. The uniform approximation has sensitivity $\kappa_\Delta>0$,
baseline $b$, error allowance $\varepsilon\geq0$, and
$C=b+\varepsilon+\log2$.
\end{theorem}

\section{Optimization Method}

\subsection{White-Box Attack}

In the white-box scenario,
we adopt the Greedy Coordinate Gradient (GCG)~\citep{zou2023universal} to optimize the adversarial prefix $a$. GCG linearizes the loss function by computing the gradient of the input embedding and identifies the best replacement candidates most likely to reduce the loss.
Specifically, for a given token $a_i$ in the prefix, we compute the linearized loss gradient of its embedding:
\begin{equation}
-\nabla_{e_{a_i}} \mathcal{L} \in \mathbb{R}^{|V|}
\end{equation}
where $e_{a_i}$ denotes the one-hot vector of the $i$-th token in the prefix, and $\nabla_{e_{a_i}} \mathcal{L}$ represents the gradient of the loss function $\mathcal{L}$ with respect to the embedding vector of token $a_i$.
Next, we select the top-$k$ tokens with the largest negative gradient values from the vocabulary to form the candidate set. Then, for each token position in the prefix, we randomly sample a batch of replacement tokens from the candidate set, substitute them, and compute the loss. The greedy algorithm then selects the candidate with the smallest loss.
This process iteratively updates each token in the prefix $a$ to minimize the overall objective function $\mathcal{L}$.

\subsection{Black-Box Attack}

Although the GCG algorithm can identify adversarial prefixes $a$, its primary limitation lies in its gradient information, rendering it unsuitable for black-box scenarios.
To overcome this limitation, 
we propose \textbf{Bayesian Prompt Optimization (BPO)}, a novel Bayesian optimization-based adversarial prompt optimization method, 
which effectively minimizes the overall objective function of resource-consuming attacks in black-box settings.
The core idea of BPO is to treat the function value feedback from each iteration as prior information, guiding the next strategy. By employing Bayesian optimization, BPO globally models the target function and efficiently explores the search space.

Bayesian optimization~\citep{frazier2018tutorial} consists of two key components: a probabilistic surrogate model, such as a Gaussian process (GP)~\cite{rasmussen2003gaussian}, which approximates the unknown objective function $-\mathcal{L}$; 
and an acquisition function $\alpha(\cdot)$, which is maximized to recommend the next query prefix by balancing exploitation and exploration 
via expected improvement in function values,
to maximize $-\mathcal{L}$.

Specifically, assume $-\mathcal{L}$ a priori follows the Gaussian process with mean $0$ and kernel function $k(\cdot,\cdot)$, denoted as $-\mathcal{L}\sim \operatorname{GP}(0, k)$. Given the observation data $\mathcal{D}=\{(\bm{a}^1, y^1),\ldots, (\bm{a}^T, y^T)\}$ where $y^i = -\mathcal{L}(\bm{a}^i)$, the predictive posterior distribution for $-\mathcal{L}(\bm{a})$ at a test prefix $\bm{a}$, denoted as $p(-\mathcal{L}(\bm{a})\mid \bm{a};\mathcal{D})$, follows a Gaussian distribution $\mathcal{N}(\mu_T(\bm{a}), \sigma_T^2(\bm{a}))$, where
\begin{align}
    \mu_T(\bm{a})&=\mathbf{k}_T(\bm{a})^\top \mathbf{K}_T^{-1} \mathbf{y}_T, \label{eq:gp-mean}\\
    \sigma_T^2(\bm{a})&=k(\bm{a}, \bm{a}) - \mathbf{k}_T(\bm{a})^\top \mathbf{K}_T^{-1} \mathbf{k}_T(\bm{a}), \label{eq:gp-variance}
\end{align}
where $\mathbf{k}_T(\bm{a})=[k(\bm{a}^1,\bm{a}),k(\bm{a}^2,\bm{a}),\ldots,k(\bm{a}^T,\bm{a})]^\top$, $\mathbf{K}_T$ 
is a $T\times T$ matrix with its $(i, j)$-th element being $k(\bm{a}^i, \bm{a}^j)$, and $\mathbf{y}_T=[y^1,\ldots,y^T]^\top$.

To address the discrete nature of the optimization search space, we design a kernel function $k(\cdot, \cdot)$ that is suitable for discrete tokens.
Specifically, we first map the discrete tokens to a continuous embedding space, obtaining $\text{emb}(a) \in \mathbb{R}^{|a| \times d}$, where $|a|$ is the number of tokens in the prefix $a$, and $d$ is the dimension of the embedding vectors. 
For two different prefixes $\bm{a}^i$ and $\bm{a}^j$, we adopt the Mat\'ern-5/2 kernel with auto-correlation in the continuous embedding space:
\begin{equation}
\begin{aligned}
  k(a^i, a^j)
  &= \sigma^2 \Bigl( 1 + \sqrt{5} \left\lVert \text{emb}(a^i) - \text{emb}(a^j) \right\rVert_L\\
  &+ \frac{5 \left\lVert \text{emb}(a^i) - \text{emb}(a^j) \right\rVert_L^2}{3} \Bigr)\\
  &\times \exp \left( -\sqrt{5} \left\lVert \text{emb}(a^i) - \text{emb}(a^j) \right\rVert_L \right),
\end{aligned}
\end{equation}
\begin{equation}
\left\lVert \text{emb}(a^i)\!-\!\text{emb}(a^j)\right\rVert_L
\!=\!\sqrt{\!\sum_{m=1}^{d}\!\frac{\left(\text{emb}(a^i)_m\!-\!\text{emb}(a^j)_m\right)^2}{\ell_m^2}}
\end{equation}
Here, $\sigma^2$ is the scale parameter of kernel, $\ell_m$ is the length scale parameter for the $m$-th feature component, representing the correlation of each dimension of the input space.

Based on the posterior distribution, we can construct the acquisition function $\alpha({\cdot})$, such as Expected Improvement (EI) \cite{movckus1975bayesian}, Probability of Improvement (PI) \cite{jones2001taxonomy}, and Upper-Confidence Bounds (UCB) \cite{srinivas2009gaussian}.
In this work, we choose UCB as the acquisition function, which is expressed as
\begin{align}
\alpha(\bm{a})=\mu_T(\bm{a})+\gamma \sigma_T(\bm{a}), \label{eq:ucb}
\end{align}
where $\gamma$ balances exploration (i.e., encouraging queries in regions with high predictive variance) and exploitation (i.e., encouraging queries in regions with high predictive mean).
Subsequently, the next point $\bm{a}^{T+1}$ is selected by maximizing the acquisition function as $\bm{a}^{T+1}=\arg\max_{\bm{a}\in A} \alpha(\bm{a})$ and then used to query the objective $-\mathcal{L}$.
Here, $A = |V|^{|a|}$ represents the search space for adversarial prefix $\bm{a}$, where $|V|$ is the vocabulary size and $|a|$ is the number of tokens in the prefix $a$.

Specifically, BPO iteratively runs the following steps to minimize the overall objective function $\mathcal{L}$: 

\textbf{Initialization}:
Randomly sample a set of prefixes $\bm{a}^i$ and obtain $y^i = -\mathcal{L}(\bm{a}^i)$ \footnote{ Since the function $\mathcal{L}_{IA}$ cannot be computed in black-box scenarios, we use the cross-entropy loss of $n$ tokens following the key reasoning tokens in relation to the short doubt-inducing phrases as a surrogate. 
} for each prefix. Construct the initial dataset $\mathcal{D} = \{(\bm{a}^i, y^i)\}_{i=1}^S$.

\textbf{Probabilistic Modeling/Model Update}:
Based on $\mathcal{D}$, model the Gaussian process $\operatorname{GP}(-\mathcal{L}, k)$ to obtain posterior mean $\mu_T(a)$ and variance estimates $\sigma_T^2(a)$ for the objective function $-\mathcal{L}(a)$, and optimize the hyperparameters.

\textbf{Acquisition}:
Construct the acquisition function $\alpha(a) = \mu_T(a) + \gamma \sigma_T(a)$. Maximize the acquisition function $a^{T+1} = \arg\max_{a \in A} \alpha(a)$ to select the next sampling prefix $a^{T+1}$.
For more efficient searching, perform $N$ steps of random replacements on $a^{T+1}$, exploring the vicinity of $a^{T+1}$. Each replacement samples a batch of replacement tokens for each token position, keeping the best replacements. The final prefix after all replacements is denoted as $a'^{T+1}$.

\textbf{Query and Iteration}:
Query the objective $-\mathcal{L}$ to obtain $y^{T+1} = -\mathcal{L}(a^{T+1})$ and $y'^{T+1} = -\mathcal{L}(a'^{T+1})$, and update the dataset: $\mathcal{D} = \mathcal{D} \cup \{(a^{T+1}, y^{T+1}), (a'^{T+1}, y'^{T+1})\}$. Repeat until the iteration limit is reached.

\section{Experiment}

\subsection{Experiment Settings}

\textbf{Datasets.} To evaluate our attack across a diverse set of reasoning scenarios, we use five benchmarks spanning math reasoning, general knowledge question answering, and code generation: GSM8K~\citep{cobbe2021training}, MathQA~\citep{amini2019mathqa}, MATH-500~\citep{lightman2023let}, MMLU-Pro~\citep{wang2024mmlu}, and HumanEval~\citep{chen2021evaluating}.
We randomly sample 100 instances from GSM8K, MathQA, MATH-500, and HumanEval, and select 10 questions from each of the 12 non-math categories in MMLU-Pro.

\textbf{Models.} We assess our attack using three state-of-the-art reasoning models: DeepSeek-R1-Distilled-Qwen-7b (R1-7b)~\citep{guo2025deepseek}, QwQ-32b~\citep{team2025qwq}, and gpt-oss-20b~\citep{agarwal2025gpt}. To simulate a realistic attack scenario, we additionally deploy a local vLLM inference server to mimic a commercial LLM service provider. More hardware and deployment details are deferred to Appendix~\ref{App:experiment_settings}.

\textbf{Baselines.} We compare our attack against the state-of-the-art resource consumption attacks that induce excessively long outputs and increase inference latency.
Specifically, we include multiple black-box baselines: chain-of-thought (CoT) prompting~\citep{wei2022chain}, CatAttack~\citep{rajeev2025cats}, and four variants of OverThink~\citep{kumar2025overthink}, including Context-Aware (CAW), Context-Agnostic (CAG), ICL-Genetic-Aware (ICLGAW), and ICL-Genetic-Agnostic (ICLGAG). For white-box baselines, we compare with the Denial-of-Service (DoS) prompt attack~\citep{geiping2024coercing}, Engorgio Prompt~\citep{dong2024engorgio}, and Excessive Reasoning Attack (ERA)~\citep{si2025excessive}.
Additionally, to provide a more direct and transparent comparison of optimization strategies, we develop three extra baselines (Heuristic Prompt, Black-box Search, Gradient-free Attack). More implementation details are provided in Appendix~\ref{App:experiment_settings}.

\textbf{Defenses.} To evaluate the performance of our attack under defenses, we select methods for reasoning compression and resource-consumption mitigation, including CCoT~\citep{cheng2024compressed}, CoD~\citep{xu2025chain}, and PD$^3$F~\citep{zhang2025pd}. We also implement and evaluate on multiple different production constraints, traditional infrastructure defenses, and post-generation defenses in a simulated server environment. More details are provided in Appendix~\ref{App:experiment_settings}.

\textbf{Metrics.} To evaluate the effectiveness of our attack, we report the widely-used metrics including average output token length (ATL), average inference time consumption (ATC), and accuracy (ACC), with the 95\% confidence intervals of ATL and ATC.
Additionally, to mimic realistic attacks against API services and assess the impact of Chatter Attack on server load and responsiveness, we also monitor the server-side runtime indicators during inference including the counts of running requests (RR), GPU cache usage (GCU), response time to first token (TTFT), and the generated token counts per second (TPS). We provide more details in Appendix~\ref{App:experiment_settings}.

\textbf{Implementation Details.} For the implementation of Chatter Attack, we use Greedy Coordinate Gradient (GCG)~\citep{zou2023universal} to optimize an adversarial prefix $a$ with a length of 20 tokens, and set the hyperparameters of the overall objective function $\mathcal{L}$ to $(\alpha, \beta, \lambda) = (1, 10, 1)$. For the implementation of Chatter-BPO Attack, we use the Adam optimizer~\citep{kingma2014adam} with a learning rate of $0.1$ to optimize the hyperparameters of the Gaussian process. We maximize the acquisition function using GCG, and then apply a dynamic random-search schedule with gradually increasing intensity to enable more efficient exploration. Additional implementation details including the steps of GCG iterations, search batch size, and specific dynamic schedule are provided in the Appendix~\ref{App:experiment_settings}.

\begin{table*}[t!]
\centering
\caption{
Quantitative results of baselines and our attacks in extending output length and latency on three datasets across three models. Chatter Attack and Chatter-BPO Attack consistently outperforms all baselines in terms of average output token length (ATL), average inference time consumption (ATC), highlighting the potential risk of resource-depletion for deployed LLM services.
}
\label{tab:main}
\resizebox{\textwidth}{!}{%
\begin{tabular}{c|l|ccc|ccc|ccc}
\toprule
\multirow{2}{*}{Model}        & \multirow{2}{*}{Method}   & \multicolumn{3}{c|}{GSM8k}                           & \multicolumn{3}{c|}{MATH-500}                       & \multicolumn{3}{c}{MathQA}                         \\
                              &                           & ATL                     & ATC                 & ACC  & ATL                    & ATC                 & ACC  & ATL                    & ATC                & ACC  \\ \midrule
\multirow{15}{*}{R1-7b}       & Base                      & 423.5 $\pm$ 16.8              & 4.7 $\pm$ 0.2             & 99.4 & 1270.6 $\pm$ 190.1           & 14.3 $\pm$ 2.2            & 92.5 & 411.1 $\pm$ 26.9             & 4.7 $\pm$ 0.3            & 52.3 \\
                              & CoT                       & 438.6 $\pm$ 27.8              & 5.1 $\pm$ 0.5             & 99.6 & 1435.7 $\pm$ 188.2           & 17.5 $\pm$ 2.1            & 96.3 & 887.4 $\pm$ 192.0            & 10.8 $\pm$ 4.2           & 54.7 \\
                              & DoS                       & 1158.7 $\pm$ 638.6            & 10.8 $\pm$ 5.0            & 96.7 & 1354.2 $\pm$ 391.2           & 14.8 $\pm$ 3.3            & 72.5 & 538.9 $\pm$ 36.3             & 6.0 $\pm$ 0.4            & 55.9 \\
                              & Engorgio Prompt           & 569.3 $\pm$ 63.2              & 6.4 $\pm$ 0.7             & 98.4 & 1524.7 $\pm$ 275.0           & 17.3 $\pm$ 3.2            & 68.6 & 798.4 $\pm$ 96.9             & 9.0 $\pm$ 1.1            & 27.6 \\
                              & CAW                       & 601.4 $\pm$ 23.8              & 6.7 $\pm$ 0.3             & 99.7 & 1089.7 $\pm$ 105.7           & 12.2 $\pm$ 1.2            & 85.7 & 445.1 $\pm$ 42.2             & 5.0 $\pm$ 0.5            & 40.1 \\
                              & CAG                       & 500.3 $\pm$ 26.3              & 5.6 $\pm$ 0.3             & 90.7 & 974.5 $\pm$ 134.5            & 11.0 $\pm$ 1.5            & 83.7 & 394.5 $\pm$ 27.6             & 4.4 $\pm$ 0.3            & 41.4 \\
                              & ICLGAW                    & 533.4 $\pm$ 29.9              & 6.0 $\pm$ 0.3             & 95.5 & 1904.2 $\pm$ 755.9           & 18.3 $\pm$ 5.9            & 63.6 & 788.5 $\pm$ 242.6            & 9.0 $\pm$ 2.8            & 56.0 \\
                              & ICLGAG                    & 851.0 $\pm$ 285.7             & 9.7 $\pm$ 3.3             & 93.3 & 861.9 $\pm$ 86.7             & 9.8 $\pm$ 1.0             & 75.3 & 788.5 $\pm$ 242.6            & 9.0 $\pm$ 2.8            & 54.7 \\
                              & CatAttack                 & 482.0 $\pm$ 21.2              & 5.5 $\pm$ 0.2             & 75.9 & 1449.8 $\pm$ 201.1           & 16.8 $\pm$ 2.3            & 86.8 & 451.8 $\pm$ 37.0             & 5.1 $\pm$ 0.4            & 36.3 \\
                              & ERA                       & 546.1 $\pm$ 29.8              & 6.5 $\pm$ 0.4             & 94.2 & 1583.3 $\pm$ 120.9           & 18.7 $\pm$ 1.4            & 91.9 & 547.5 $\pm$ 63.7             & 5.8 $\pm$ 0.7            & 35.6 \\
                              & Gradient-free Attack      & 474.9 $\pm$ 32.8              & 6.0 $\pm$ 0.4             & 88.6 & 1317.5 $\pm$ 130.9           & 16.7 $\pm$ 1.6            & 60.2 & 547.3 $\pm$ 57.5             & 6.5 $\pm$ 0.7            & 40.0 \\
                              & Black-box Search          & 1456.7 $\pm$ 594.6            & 14.7 $\pm$ 4.3            & 84.9 & 2357.2 $\pm$ 237.1           & 28.0 $\pm$ 2.9            & 87.9 & 942.3 $\pm$ 51.5             & 11.1 $\pm$ 0.6           & 54.9 \\
                              & Heuristic Prompt          & 883.5 $\pm$ 54.1              & 10.6 $\pm$ 0.7            & 78.1 & 1794.0 $\pm$ 167.1           & 21.6 $\pm$ 2.0            & 91.0 & 1156.1 $\pm$ 67.7            & 13.9 $\pm$ 0.8           & 54.4 \\
                              & \cellcolor{gray!20}\textbf{Chatter Attack (ours)}     & \cellcolor{gray!20}\textbf{13345.8 $\pm$ 1067.4} & \cellcolor{gray!20}\textbf{105.4 $\pm$ 16.4} & \cellcolor{gray!20}70.4 & \cellcolor{gray!20}\textbf{9577.8 $\pm$ 1329.4} & \cellcolor{gray!20}\textbf{81.7 $\pm$ 8.9}   & \cellcolor{gray!20}16.4 & \cellcolor{gray!20}\textbf{7429.2 $\pm$ 1312.1} & \cellcolor{gray!20}\ul{67.3 $\pm$ 9.5}     & \cellcolor{gray!20}27.3 \\
                              & \cellcolor{gray!20}\textbf{Chatter-BPO Attack (ours)} & \cellcolor{gray!20}\ul{7846.9 $\pm$ 1473.5}     & \cellcolor{gray!20}\ul{91.7 $\pm$ 17.4}     & \cellcolor{gray!20}67.6 & \cellcolor{gray!20}\ul{6660.6 $\pm$ 1464.9}    & \cellcolor{gray!20}\ul{77.7 $\pm$ 17.3}     & \cellcolor{gray!20}13.5 & \cellcolor{gray!20}\ul{7338.2 $\pm$ 1455.6}    & \cellcolor{gray!20}\textbf{87.5 $\pm$ 17.1} & \cellcolor{gray!20}22.9 \\ \midrule
\multirow{15}{*}{QwQ-32b}     & Base                      & 991.4 $\pm$ 149.4             & 25.0 $\pm$ 3.7            & 99.5 & 3007.0 $\pm$ 406.4           & 68.5 $\pm$ 7.6            & 94.4 & 3955.6 $\pm$ 243.2           & 95.4 $\pm$ 4.6           & 55.8 \\
                              & CoT                       & 1016.5 $\pm$ 137.1            & 25.5 $\pm$ 3.5            & 95.7 & 3102.5 $\pm$ 488.8           & 69.5 $\pm$ 8.0            & 97.8 & 3805.9 $\pm$ 230.8           & 92.5 $\pm$ 4.5           & 66.4 \\
                              & DoS                       & 828.6 $\pm$ 82.0              & 20.9 $\pm$ 2.1            & 99.3 & 2600.3 $\pm$ 351.2           & 62.8 $\pm$ 7.7            & 96.7 & 3159.4 $\pm$ 249.6           & 78.6 $\pm$ 5.8           & 55.6 \\
                              & Engorgio Prompt           & 1450.2 $\pm$ 120.8            & 38.2 $\pm$ 3.2            & 99.6 & 3964.5 $\pm$ 400.7           & 90.4 $\pm$ 6.6            & 92.6 & \ul{4980.3 $\pm$ 750.5}           & 107.1 $\pm$ 19.1         & 45.4 \\
                              & CAW                       & 2314.0 $\pm$ 337.9            & 56.5 $\pm$ 7.7            & 99.3 & 2973.9 $\pm$ 432.9           & 67.4 $\pm$ 8.0            & 93.3 & 2872.4 $\pm$ 266.2           & 71.0 $\pm$ 5.9           & 44.5 \\
                              & CAG                       & 806.6 $\pm$ 97.5              & 20.4 $\pm$ 2.5            & 98.3 & 2235.5 $\pm$ 413.9           & 53.0 $\pm$ 6.9            & 92.2 & 2093.5 $\pm$ 209.8           & 52.9 $\pm$ 5.2           & 46.6 \\
                              & ICLGAW                    & 2275.5 $\pm$ 251.6            & 58.5 $\pm$ 6.6            & 97.7 & 4266.6 $\pm$ 453.8           & 91.1 $\pm$ 7.7            & 93.7 & 3885.8 $\pm$ 453.3           & 85.2 $\pm$ 12.1          & 45.4 \\
                              & ICLGAG                    & 1509.2 $\pm$ 283.4            & 39.8 $\pm$ 7.4            & 98.9 & 3893.3 $\pm$ 857.2           & 95.2 $\pm$ 21.3           & 87.9 & 3247.7 $\pm$ 356.6           & 58.9 $\pm$ 9.4           & 51.0 \\
                              & CatAttack                 & 1756.0 $\pm$ 206.1            & 44.2 $\pm$ 5.3            & 99.6 & 2741.1 $\pm$ 317.5           & 66.9 $\pm$ 7.1            & 90.7 & 3483.3 $\pm$ 237.3           & 84.9 $\pm$ 5.0           & 52.1 \\
                              & ERA                       & 2747.9 $\pm$ 284.0            & 65.7 $\pm$ 6.7            & 76.4 & 2875.4 $\pm$ 415.7           & 63.9 $\pm$ 6.6            & 92.4 & 3931.5 $\pm$ 234.8           & 91.8 $\pm$ 4.4           & 51.3 \\
                              & Gradient-free Attack      & 914.1 $\pm$ 115.0             & 23.0 $\pm$ 2.9            & 99.8 & 3121.7 $\pm$ 408.5           & 69.3 $\pm$ 7.0            & 94.4 & 3989.9 $\pm$ 228.3           & 96.5 $\pm$ 4.4           & 52.3 \\
                              & Black-box Search          & 2573.6 $\pm$ 433.9            & 60.2 $\pm$ 8.2            & 98.2 & 2600.4 $\pm$ 245.6           & 65.9 $\pm$ 6.3            & 92.2 & 4038.0 $\pm$ 301.4           & 102.6 $\pm$ 7.8          & 54.3 \\
                              & Heuristic Prompt          & 2634.0 $\pm$ 388.6            & 61.5 $\pm$ 7.8            & 99.3 & 3687.8 $\pm$ 359.9           & 86.4 $\pm$ 6.9            & 96.9 & 3692.1 $\pm$ 341.9           & 85.3 $\pm$ 4.4           & 51.9 \\
                              & \cellcolor{gray!20}\textbf{Chatter Attack (ours)}     & \cellcolor{gray!20}\textbf{3298.4 $\pm$ 370.7}   & \cellcolor{gray!20}\ul{77.3 $\pm$ 7.3}      & \cellcolor{gray!20}76.3 & \cellcolor{gray!20}\textbf{5303.3 $\pm$ 696.3}  & \cellcolor{gray!20}\textbf{122.1 $\pm$ 11.5} & \cellcolor{gray!20}61.1 & \cellcolor{gray!20}\textbf{6915.4 $\pm$ 615.7}  & \cellcolor{gray!20}\textbf{112.0 $\pm$ 3.8} & \cellcolor{gray!20}52.2 \\
                              & \cellcolor{gray!20}\textbf{Chatter-BPO Attack (ours)} & \cellcolor{gray!20}\ul{3149.8 $\pm$ 489.1}      & \cellcolor{gray!20}\textbf{80.6 $\pm$ 12.7}  & \cellcolor{gray!20}66.0 & \cellcolor{gray!20}\ul{4800.0 $\pm$ 450.3}     & \cellcolor{gray!20}\ul{105.8 $\pm$ 10.3}    & \cellcolor{gray!20}55.5 & \cellcolor{gray!20}4811.8 $\pm$ 296.1     & \cellcolor{gray!20}\ul{109.1 $\pm$ 3.6}    & \cellcolor{gray!20}45.6 \\ \midrule
\multirow{15}{*}{gpt-oss-20b} & Base                      & 253.0 $\pm$ 35.5              & 1.3 $\pm$ 0.2             & 97.8 & 389.9 $\pm$ 53.5             & 2.0 $\pm$ 0.3             & 81.3 & 112.8 $\pm$ 10.8             & 0.6 $\pm$ 0.1            & 43.1 \\
                              & CoT                       & 420.3 $\pm$ 56.0              & 2.2 $\pm$ 0.3             & 96.7 & 558.2 $\pm$ 68.4             & 2.9 $\pm$ 0.4             & 76.9 & 240.4 $\pm$ 14.5             & 1.2 $\pm$ 0.1            & 38.2 \\
                              & DoS                       & 363.8 $\pm$ 80.1              & 2.0 $\pm$ 0.4             & 86.7 & 850.2 $\pm$ 80.4             & 4.5 $\pm$ 0.6             & 87.9 & 478.6 $\pm$ 33.5             & 2.4 $\pm$ 0.2            & 45.2 \\
                              & Engorgio Prompt           & 850.3 $\pm$ 280.6             & 5.3 $\pm$ 2.0             & 97.7 & 826.6 $\pm$ 70.4             & 3.8 $\pm$ 0.5             & 72.2 & 400.8 $\pm$ 43.5             & 2.5 $\pm$ 0.3            & 44.5 \\
                              & CAW                       & 248.9 $\pm$ 26.6              & 1.4 $\pm$ 0.2             & 97.5 & 360.4 $\pm$ 41.6             & 1.9 $\pm$ 0.2             & 85.6 & 192.3 $\pm$ 23.4             & 1.0 $\pm$ 0.1            & 20.9 \\
                              & CAG                       & 201.2 $\pm$ 19.8              & 1.0 $\pm$ 0.1             & 98.3 & 327.1 $\pm$ 43.1             & 1.7 $\pm$ 0.2             & 88.5 & 136.3 $\pm$ 9.8              & 0.7 $\pm$ 0.1            & 21.1 \\
                              & ICLGAW                    & 859.2 $\pm$ 156.8             & 5.2 $\pm$ 0.9             & 97.2 & 1258.9 $\pm$ 285.4           & 6.6 $\pm$ 1.5             & 92.8 & 784.7 $\pm$ 45.6             & 3.2 $\pm$ 0.3            & 47.8 \\
                              & ICLGAG                    & 824.6 $\pm$ 123.7             & 4.3 $\pm$ 0.6             & 98.1 & 1156.8 $\pm$ 258.9           & 6.0 $\pm$ 1.4             & 87.7 & 487.3 $\pm$ 48.1             & 2.2 $\pm$ 0.2            & 43.4 \\
                              & CatAttack                 & 463.6 $\pm$ 53.8              & 2.4 $\pm$ 0.3             & 96.7 & 594.6 $\pm$ 54.5             & 3.1 $\pm$ 0.3             & 96.2 & 260.5 $\pm$ 14.1             & 1.5 $\pm$ 0.1            & 30.0 \\
                              & ERA                       & 420.0 $\pm$ 56.2              & 2.5 $\pm$ 0.4             & 98.9 & 358.7 $\pm$ 33.1             & 1.4 $\pm$ 0.1             & 87.8 & 322.0 $\pm$ 30.5             & 1.2 $\pm$ 0.1            & 22.3 \\
                              & Gradient-free Attack      & 372.0 $\pm$ 27.0              & 1.4 $\pm$ 0.1             & 96.4 & 562.9 $\pm$ 67.9             & 2.2 $\pm$ 0.3             & 84.7 & 247.0 $\pm$ 12.4             & 0.9 $\pm$ 0.0            & 45.1 \\
                              & Black-box Search          & 504.2 $\pm$ 64.6              & 4.6 $\pm$ 0.6             & 99.5 & 658.8 $\pm$ 61.1             & 3.6 $\pm$ 0.3             & 87.9 & 300.8 $\pm$ 18.7             & 2.7 $\pm$ 0.2            & 42.4 \\
                              & Heuristic Prompt          & 1055.2 $\pm$ 374.8            & 4.2 $\pm$ 1.6             & 95.6 & 936.9 $\pm$ 88.1             & 6.4 $\pm$ 0.6             & 85.3 & 584.2 $\pm$ 51.9             & 2.2 $\pm$ 0.2            & 35.9 \\
                              & \cellcolor{gray!20}\textbf{Chatter Attack (ours)}     & \cellcolor{gray!20}\textbf{1980.3 $\pm$ 508.8}   & \cellcolor{gray!20}\textbf{11.4 $\pm$ 3.3}   & \cellcolor{gray!20}71.2 & \cellcolor{gray!20}\ul{1893.0 $\pm$ 332.7}     & \cellcolor{gray!20}\ul{9.6 $\pm$ 1.4}       & \cellcolor{gray!20}52.0 & \cellcolor{gray!20}\textbf{910.5 $\pm$ 139.8}   & \cellcolor{gray!20}\textbf{5.1 $\pm$ 0.8}   & \cellcolor{gray!20}14.4 \\
                              & \cellcolor{gray!20}\textbf{Chatter-BPO Attack (ours)} & \cellcolor{gray!20}\ul{1182.7 $\pm$ 172.5}      & \cellcolor{gray!20}\ul{6.6 $\pm$ 0.7}       & \cellcolor{gray!20}83.0 & \cellcolor{gray!20}\textbf{2438.7 $\pm$ 344.3}  & \cellcolor{gray!20}\textbf{10.6 $\pm$ 1.9}   & \cellcolor{gray!20}56.7 & \cellcolor{gray!20}\ul{820.5 $\pm$ 51.8}       & \cellcolor{gray!20}\ul{4.2 $\pm$ 1.2}      & \cellcolor{gray!20}18.2 \\ \bottomrule
\end{tabular}
}
\vspace{-2ex}
\end{table*}

\subsection{Main Results}

We demonstrate the effectiveness of our Chatter Attack and Chatter-BPO Attack in inducing long responses across multiple models and datasets. As shown in Table~\ref{tab:main}, Chatter Attack consistently achieves the largest average output token length (ATL) and the highest average inference time consumption (ATC) among all baselines, indicating attack effectiveness; in particular, on R1-7b it increases ATL by up to $31.5\times$ compared to benign generation for GSM8K. Chatter-BPO Attack achieves the second-best performance, while outperforming all baseline methods.
We present the optimized attack prefixes and corresponding output examples in Table~\ref{example_7b}. The results show that both Chatter Attack and Chatter-BPO Attack successfully induce the target skeptical phrases, which then trigger substantially longer generations than the benign outputs. Meanwhile, we observe a drop in accuracy (ACC) in some settings, mainly because overly long outputs often hit the generation budget and fail to return a valid final answer. Overall, these results highlight the efficacy of our attacks.

\subsection{Generalization to Different Reasoning Settings}

In addition to mathematical reasoning, our attacks could generalize to broader reasoning settings, for example question-answering or code generation. As shown in Table~\ref{tab:other_reasoning}, our attacks remain effective on QA benchmarks such as MMLU-Pro and coding benchmarks such as HumanEval, consistently achieving superior ATL and ATC compared to the baselines, which shows strong generalizability.

\subsection{Attack Results against Defenses}

We then study the impact of defenses on our attacks. We consider several representative defenses, including prompt-based methods (CCoT and CoD) and decoding-strategy defense (PD$^3$F). For comparison, we report the performance of DoS under the same defenses.
In Table~\ref{tab:defense}, although these defenses reduce the ATL and ATC induced by our attacks to some extent, it is worth noting that none of them can fully mitigate the impact: even with defenses enabled, Chatter and Chatter-BPO Attack still lead to a significant increase in ATL and ATC, highlighting their efficacy against mitigation. More results against multiple different production constraints, traditional infrastructure defenses, and post-generation defense are provided in Appendix~\ref{app:defense}.

\subsection{Attack Results on LLM Service}

\begin{table*}[t]
\centering
\caption{
Quantitative results of DoS attack and our attacks against multiple defenses on three datasets.
Although the performance of Chatter Attack and Chatter-BPO Attack decrease when facing defenses, none of the defense can fully mitigate the impact.
}
\label{tab:defense}
\resizebox{\textwidth}{!}{%
\begin{tabular}{c|l|ccc|ccc|ccc}
\toprule
\multirow{2}{*}{Attack}        & \multirow{2}{*}{Defense}   & \multicolumn{3}{c|}{GSM8k}                           & \multicolumn{3}{c|}{MATH-500}                       & \multicolumn{3}{c}{MathQA}                         \\

                                      &            & ATL            & ATC        & ACC  & ATL           & ATC       & ACC  & ATL           & ATC       & ACC  \\ \midrule
No Attack                                  & No Defense & 423.5 $\pm$ 16.8     & 4.7 $\pm$ 0.2    & 99.4 & 1270.6 $\pm$ 190.1  & 14.3 $\pm$ 2.2  & 92.5 & 411.1 $\pm$ 26.9    & 4.7 $\pm$ 0.3   & 52.3 \\ \midrule
\multirow{4}{*}{DoS}                  & No Defense & 1158.7 $\pm$ 638.6   & 10.8 $\pm$ 5.0   & 96.7 & 1354.2 $\pm$ 391.2  & 14.8 $\pm$ 3.3  & 72.5 & 538.9 $\pm$ 36.3    & 6.0 $\pm$ 0.4   & 55.9 \\
                                      & CCoT       & 950.2 $\pm$ 457.8    & 9.2 $\pm$ 4.1    & 88.5 & 1226.6 $\pm$ 354.5  & 13.1 $\pm$ 2.9  & 65.2 & 486.5 $\pm$ 31.0    & 5.3 $\pm$ 0.3   & 48.7 \\
                                      & CoD        & 857.7 $\pm$ 385.6    & 8.1 $\pm$ 3.5    & 92.3 & 1167.4 $\pm$ 396.8  & 12.0 $\pm$ 2.5  & 68.9 & 420.8 $\pm$ 25.7    & 4.6 $\pm$ 0.3   & 52.4 \\
                                      & PD$^3$F    & 1051.5 $\pm$ 523.3   & 9.8 $\pm$ 4.6    & 90.1 & 1253.3 $\pm$ 378.4  & 13.6 $\pm$ 3.1  & 70.8 & 518.6 $\pm$ 33.8    & 5.6 $\pm$ 0.4   & 50.2 \\ \midrule
\multirow{4}{*}{\textbf{Chatter   Attack (ours)}}     & No Defense & 13345.8 $\pm$ 1067.4 & 105.4 $\pm$ 16.4 & 70.4 & 9577.8 $\pm$ 1329.4 & 81.7 $\pm$ 8.9  & 16.4 & 7429.2 $\pm$ 1312.1 & 67.3 $\pm$ 9.5  & 27.3 \\
                                      & CCoT       & 7808.9 $\pm$ 1943.7  & 61.3 $\pm$ 13.8  & 25.3 & 8281.3 $\pm$ 1336.3 & 72.8 $\pm$ 9.3  & 6.8  & 3913.0 $\pm$ 1211.0 & 34.2 $\pm$ 8.8  & 15.2 \\
                                      & CoD        & 5088.4 $\pm$ 1352.1  & 43.2 $\pm$ 9.7   & 34.6 & 4819.2 $\pm$ 1187.9 & 44.7 $\pm$ 9.0  & 7.3  & 3330.6 $\pm$ 1191.6 & 28.1 $\pm$ 8.6  & 24.5 \\
                                      & PD$^3$F    & 6698.3 $\pm$ 1764.2  & 55.9 $\pm$ 11.4  & 55.9 & 5092.0 $\pm$ 1305.2 & 46.2 $\pm$ 9.2  & 14.9 & 4528.9 $\pm$ 1531.8 & 40.5 $\pm$ 9.0  & 22.4 \\ \midrule
\multirow{4}{*}{\textbf{Chatter-BPO   Attack (ours)}} & No Defense & 7846.9 $\pm$ 1473.5  & 91.7 $\pm$ 17.4  & 67.6 & 6660.6 $\pm$ 1464.9 & 77.7 $\pm$ 17.3 & 13.5 & 7338.2 $\pm$ 1455.6 & 87.5 $\pm$ 17.1 & 22.9 \\
                                      & CCoT       & 5579.2 $\pm$ 1243.5  & 52.3 $\pm$ 10.8  & 40.2 & 5890.5 $\pm$ 1103.8 & 53.4 $\pm$ 9.7  & 10.5 & 3226.8 $\pm$ 981.6  & 30.2 $\pm$ 7.8  & 18.3 \\
                                      & CoD        & 4202.6 $\pm$ 951.7   & 38.9 $\pm$ 8.2   & 45.8 & 3993.7 $\pm$ 859.4  & 36.2 $\pm$ 7.5  & 12.1 & 2884.4 $\pm$ 752.9  & 25.4 $\pm$ 6.3  & 28.7 \\
                                      & PD$^3$F    & 6395.8 $\pm$ 1340.3  & 48.7 $\pm$ 9.5   & 50.3 & 4554.2 $\pm$ 958.1  & 41.8 $\pm$ 8.2  & 16.8 & 3817.6 $\pm$ 1127.2 & 35.2 $\pm$ 8.4  & 25.6 \\ \bottomrule
\end{tabular}
}
\vspace{-2ex}
\end{table*}

\begin{figure*}[t!]
    \centering
    \includegraphics[width=\textwidth]{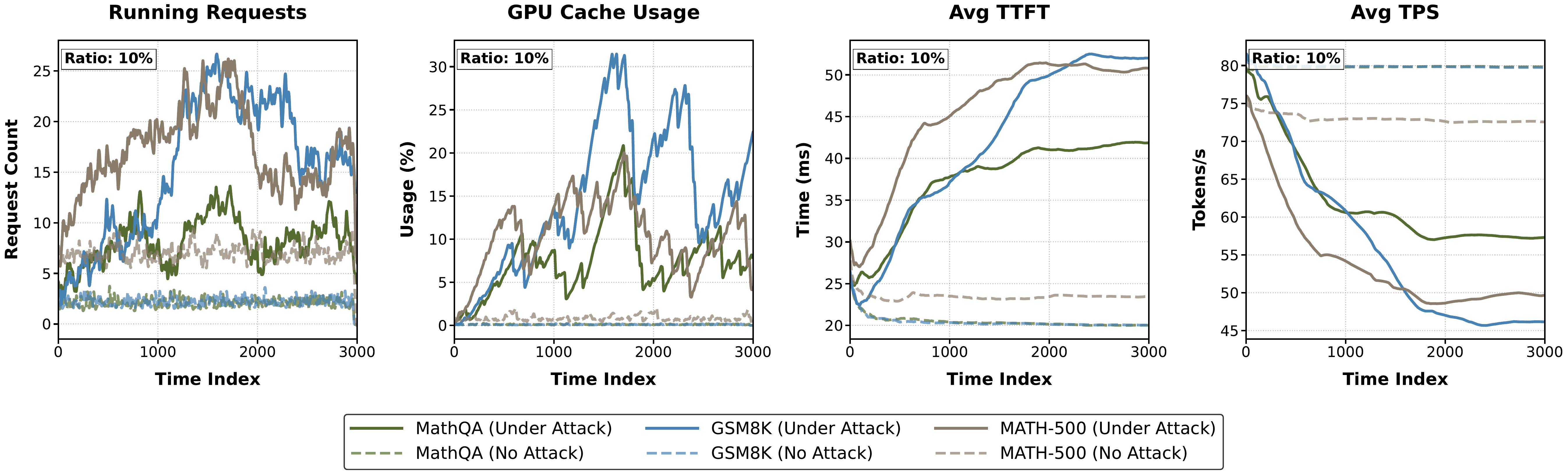}
    \vspace{-3ex}
    \caption{
        Server-side runtime metrics of the deployed LLM service under purely benign traffic and mixed traffic containing 10\% Chatter Attack queries. The service is evaluated at a constant rate of 20 requests per minute. Compared to the benign baseline, injecting just 10\% attack queries leads to a significantly larger backlog of pending requests and higher GPU cache usage, resulting in the slow response compared to benign queries.
    }
    \vspace{-3ex}
    \label{fig4}
\end{figure*}

To more realistically emulate an LLM service operator, we deploy an R1-7B inference service and simulate an attack by sending a mix of benign and Chatter Attack queries at a fixed rate (e.g., 20 requests per minute). As shown in Figure~\ref{fig4}, compared to purely benign traffic, requests injected with only 10\% Chatter Attack queries lead to a large backlog of pending requests, significantly increasing GPU cache usage and placing a substantial burden on the server.
Moreover, these requests slow down the service response: the time to first token (TTFT) increases by $2\times$ to $3\times$, and the generation throughput drops from about 80 tokens/s to lower than 60 tokens/s, severely hindering normal service operation. Furthermore, we evaluate the impact of varying attack injection rates in Appendix~\ref{app:service}, demonstrating that even a marginal 1\% of attack traffic is sufficient to cause noticeable fluctuations in response latency. Overall, these results provide a realistic demonstration of the potential risks and system-level impacts of resource exhaustion attacks in practice.

\subsection{Attack Results against Production Constraints and Infrastructure Defenses.}

\begin{wraptable}{r}{0.48\textwidth}
  \centering
  \small
  \setlength{\tabcolsep}{2pt}
  \vspace{-3ex}
  \caption{Chatter-BPO Attack under multiple service-side defenses in an
  emulated deployment. Although some defenses reduce the impact of the attack,
  service metrics do not fully recover to the benign baseline.}
  \label{tab:service_defenses_main}

  \resizebox{\linewidth}{!}{%
  \begin{tabular}{lrrrr}
    \toprule
    Setting & RR & GCU (\%) & TTFT (ms) & TPS \\
    \midrule
    Benign & 2.2 & 0.1 & 22.4 & 78.8 \\
    Under Attack & 17.4 & 17.3 & 40.1 & 60.3 \\
    \midrule
    Output-Token Limit & 6.2 & 3.0 & 30.0 & 71.3 \\
    Per-User Rate Limits & 16.4 & 16.3 & 42.9 & 56.0 \\
    IP Blocking (Single IP) & 7.4 & 7.6 & 36.8 & 60.1 \\
    IP Blocking (IP Rotation) & 19.3 & 18.7 & 43.8 & 52.9 \\
    Output Loop Detection & 9.1 & 9.2 & 39.6 & 58.6 \\
    Excessive-Reasoning Monitoring & 6.2 & 5.9 & 31.4 & 67.5 \\
    \bottomrule
  \end{tabular}%
  }
\end{wraptable}

In real-world deployments, API services commonly operate under multiple production constraints and are equipped with various defenses and output-side monitoring. To more comprehensively evaluate Chatter Attack, we implement similar defenses in a simulated LLM service. As shown in Table~\ref{tab:service_defenses_main}, directly limiting output length substantially mitigates the attack, also results in 6.2 RR, roughly three times the benign baseline. User rate limits provide only limited improvements to the resource exhaustion. IP blocking offers some mitigation when the attacker uses a fixed identity, but RR rises to 19.3 when the attacker rotates IP addresses. Output loop detection and excessive-reasoning monitoring likewise reduce RR, but service metrics still fail to recover to the benign baseline. Overall, these defenses mitigate the attack to varying degrees, but none can fully eliminate the impact of Chatter Attack. Implementation settings and more results under additional defenses are provided in Appendices~\ref{App:experiment_settings} and~\ref{app:defense}, respectively.

\subsection{Ablation Studies, Sensitivity Analysis and Additional Experiments}

We conduct detailed ablation studies on the three terms in the Chatter Attack objective. As shown in Table~\ref{tab:loss_ablation}, the full objective achieves the best performance, while removing either $\mathcal{L}_{\text{IA}}$ or $\mathcal{L}_{\text{GE}}$ leads to a consistent degradation. More detailed results and analyses are provided in Appendix~\ref{app:ablation}.
We also report hyperparameter sensitivity analyses for the objective (Appendix~\ref{app:sensitive}), the sensitivity of the attack prefix length (Appendix~\ref{app:prefix_length}), and an ablation study of the Chatter-BPO Attack (Appendix~\ref{app:bpo_sensitive}). Additionally, we evaluate the robustness to decoding strategies (Appendix~\ref{app:decoding}) and quantization (Appendix~\ref{app:quantization}), provide the full attack results on deployed LLM services (Appendix~\ref{app:service}), and report the attack effectiveness and transferability on real API services (Appendix~\ref{app:api}) and resilience against additional deployment constraints and defenses (Appendix~\ref{app:defense}).

\vspace{-1ex}
\section{Conclusion}
\vspace{-1ex}

In this paper, we introduce the Chatter Attack and Bayesian Prompt Optimization (BPO) as effective strategies to exploit the resource consumption vulnerabilities of LLMs. By optimizing short target phrases to induce self-doubt and maximize attention and entropy during the generation process, the Chatter Attack successfully manipulates output length, even in black-box settings with BPO.
We further derive a lower bound on the expected generation length in terms of the three-loss surrogate objective, providing theoretical support for our loss design.
Experiments demonstrate the superiority of our approach, with significant increases in response length and evidence of potential resource depletion and service disruption. 
These findings underscore the importance of developing robust defensive mechanisms to protect against such adversarial attacks in real-world LLM deployments.

\textbf{Limitation.} A limitation of our work is that the proposed attack relies on the sequential decoding process of autoregressive LLMs and cannot be directly applied to diffusion language models, which generate tokens non-sequentially. Moreover, the attack is less effective on conventional conversational models without explicit chain-of-thought reasoning, requiring a redesign of the target phrases and the set of critical tokens. We leave extensions to non-reasoning conversational models and non-autoregressive architectures for future research.

\begin{ack}
This paper is supported by the Strategic Priority Research Program of CAS Grant No. XDA0480502 and XDA0480503, the NSFC Projects (62276149, U2341228, 62595771, 62676212) and Beijing Nova Program (202604841276). Yibo Miao is also supported by the Young Elite Scientists Sponsorship Program of the China Association for Science and Technology (CAST).
\end{ack}

\bibliography{example_paper}

@article{zhao2023survey,
  title={A survey of large language models},
  author={Zhao, Wayne Xin and Zhou, Kun and Li, Junyi and Tang, Tianyi and Wang, Xiaolei and Hou, Yupeng and Min, Yingqian and Zhang, Beichen and Zhang, Junjie and Dong, Zican and others},
  journal={arXiv preprint arXiv:2303.18223},
  volume={1},
  number={2},
  year={2023}
}

@article{chang2024survey,
  title={A survey on evaluation of large language models},
  author={Chang, Yupeng and Wang, Xu and Wang, Jindong and Wu, Yuan and Yang, Linyi and Zhu, Kaijie and Chen, Hao and Yi, Xiaoyuan and Wang, Cunxiang and Wang, Yidong and others},
  journal={ACM transactions on intelligent systems and technology},
  volume={15},
  number={3},
  pages={1--45},
  year={2024},
  publisher={ACM New York, NY}
}

@article{naveed2025comprehensive,
  title={A comprehensive overview of large language models},
  author={Naveed, Humza and Khan, Asad Ullah and Qiu, Shi and Saqib, Muhammad and Anwar, Saeed and Usman, Muhammad and Akhtar, Naveed and Barnes, Nick and Mian, Ajmal},
  journal={ACM Transactions on Intelligent Systems and Technology},
  volume={16},
  number={5},
  pages={1--72},
  year={2025},
  publisher={ACM New York, NY}
}

@article{wei2022chain,
  title={Chain-of-thought prompting elicits reasoning in large language models},
  author={Wei, Jason and Wang, Xuezhi and Schuurmans, Dale and Bosma, Maarten and Xia, Fei and Chi, Ed and Le, Quoc V and Zhou, Denny and others},
  journal={Advances in neural information processing systems},
  volume={35},
  pages={24824--24837},
  year={2022}
}

@article{jaech2024openai,
  title={Openai o1 system card},
  author={Jaech, Aaron and Kalai, Adam and Lerer, Adam and Richardson, Adam and El-Kishky, Ahmed and Low, Aiden and Helyar, Alec and Madry, Aleksander and Beutel, Alex and Carney, Alex and others},
  journal={arXiv preprint arXiv:2412.16720},
  year={2024}
}

@article{guo2025deepseek,
  title={Deepseek-r1: Incentivizing reasoning capability in llms via reinforcement learning},
  author={Guo, Daya and Yang, Dejian and Zhang, Haowei and Song, Junxiao and Zhang, Ruoyu and Xu, Runxin and Zhu, Qihao and Ma, Shirong and Wang, Peiyi and Bi, Xiao and others},
  journal={arXiv preprint arXiv:2501.12948},
  year={2025}
}

@misc{team2025qwq,
  title={Qwq-32b: Embracing the power of reinforcement learning},
  author={Team, Qwen},
  year={2025},
  publisher={March}
}

@article{feng2023towards,
  title={Towards revealing the mystery behind chain of thought: a theoretical perspective},
  author={Feng, Guhao and Zhang, Bohang and Gu, Yuntian and Ye, Haotian and He, Di and Wang, Liwei},
  journal={Advances in Neural Information Processing Systems},
  volume={36},
  pages={70757--70798},
  year={2023}
}

@article{wu2025more,
  title={When more is less: Understanding chain-of-thought length in llms},
  author={Wu, Yuyang and Wang, Yifei and Ye, Ziyu and Du, Tianqi and Jegelka, Stefanie and Wang, Yisen},
  journal={arXiv preprint arXiv:2502.07266},
  year={2025}
}

@article{ahn2024large,
  title={Large language models for mathematical reasoning: Progresses and challenges},
  author={Ahn, Janice and Verma, Rishu and Lou, Renze and Liu, Di and Zhang, Rui and Yin, Wenpeng},
  journal={arXiv preprint arXiv:2402.00157},
  year={2024}
}

@article{yang2023coupling,
  title={Coupling large language models with logic programming for robust and general reasoning from text},
  author={Yang, Zhun and Ishay, Adam and Lee, Joohyung},
  journal={arXiv preprint arXiv:2307.07696},
  year={2023}
}

@article{liu2025logical,
  title={Logical reasoning in large language models: A survey},
  author={Liu, Hanmeng and Fu, Zhizhang and Ding, Mengru and Ning, Ruoxi and Zhang, Chaoli and Liu, Xiaozhang and Zhang, Yue},
  journal={arXiv preprint arXiv:2502.09100},
  year={2025}
}

@article{patterson2021carbon,
  title={Carbon emissions and large neural network training},
  author={Patterson, David and Gonzalez, Joseph and Le, Quoc and Liang, Chen and Munguia, Lluis-Miquel and Rothchild, Daniel and So, David and Texier, Maud and Dean, Jeff},
  journal={arXiv preprint arXiv:2104.10350},
  year={2021}
}

@article{patterson2022carbon,
  title={The carbon footprint of machine learning training will plateau, then shrink},
  author={Patterson, David and Gonzalez, Joseph and H{\"o}lzle, Urs and Le, Quoc and Liang, Chen and Munguia, Lluis-Miquel and Rothchild, Daniel and So, David R and Texier, Maud and Dean, Jeff},
  journal={Computer},
  volume={55},
  number={7},
  pages={18--28},
  year={2022},
  publisher={IEEE}
}

@inproceedings{patel2024splitwise,
  title={Splitwise: Efficient generative llm inference using phase splitting},
  author={Patel, Pratyush and Choukse, Esha and Zhang, Chaojie and Shah, Aashaka and Goiri, {\'I}{\~n}igo and Maleki, Saeed and Bianchini, Ricardo},
  booktitle={2024 ACM/IEEE 51st Annual International Symposium on Computer Architecture (ISCA)},
  pages={118--132},
  year={2024},
  organization={IEEE}
}

@inproceedings{li2024llm,
  title={Llm inference serving: Survey of recent advances and opportunities},
  author={Li, Baolin and Jiang, Yankai and Gadepally, Vijay and Tiwari, Devesh},
  booktitle={2024 IEEE High Performance Extreme Computing Conference (HPEC)},
  pages={1--8},
  year={2024},
  organization={IEEE}
}

@article{gao2024denial,
  title={Denial-of-service poisoning attacks against large language models},
  author={Gao, Kuofeng and Pang, Tianyu and Du, Chao and Yang, Yong and Xia, Shu-Tao and Lin, Min},
  journal={arXiv preprint arXiv:2410.10760},
  year={2024}
}

@article{geiping2024coercing,
  title={Coercing llms to do and reveal (almost) anything},
  author={Geiping, Jonas and Stein, Alex and Shu, Manli and Saifullah, Khalid and Wen, Yuxin and Goldstein, Tom},
  journal={arXiv preprint arXiv:2402.14020},
  year={2024}
}

@inproceedings{shumailov2021sponge,
  title={Sponge examples: Energy-latency attacks on neural networks},
  author={Shumailov, Ilia and Zhao, Yiren and Bates, Daniel and Papernot, Nicolas and Mullins, Robert and Anderson, Ross},
  booktitle={2021 IEEE European symposium on security and privacy (EuroS\&P)},
  pages={212--231},
  year={2021},
  organization={IEEE}
}

@inproceedings{chen2022nicgslowdown,
  title={Nicgslowdown: Evaluating the efficiency robustness of neural image caption generation models},
  author={Chen, Simin and Song, Zihe and Haque, Mirazul and Liu, Cong and Yang, Wei},
  booktitle={Proceedings of the IEEE/CVF Conference on Computer Vision and Pattern Recognition},
  pages={15365--15374},
  year={2022}
}

@inproceedings{dong2025engorgio,
  title={An Engorgio Prompt Makes Large Language Model Babble on},
  author={Dong, Jianshuo and Zhang, Ziyuan and Zhang, Qingjie and Zhang, Tianwei and Wang, Hao and Li, Hewu and Li, Qi and Zhang, Chao and Xu, Ke and Qiu, Han},
  booktitle={International Conference on Learning Representations},
  year={2025}
}

@article{si2025excessive,
  title={Excessive Reasoning Attack on Reasoning LLMs},
  author={Si, Wai Man and Li, Mingjie and Backes, Michael and Zhang, Yang},
  journal={arXiv preprint arXiv:2506.14374},
  year={2025}
}

@article{yan2025bithydra,
  title={Bithydra: Towards bit-flip inference cost attack against large language models},
  author={Yan, Xiaobei and Li, Yiming and Wang, Hao and Qiu, Han and Zhang, Tianwei},
  journal={arXiv preprint arXiv:2505.16670},
  year={2025}
}

@inproceedings{gao2024inducing,
  title={Inducing High Energy-Latency of Large Vision-Language Models with Verbose Images},
  author={Gao, Kuofeng and Bai, Yang and Gu, Jindong and Xia, Shu-Tao and Torr, Philip and Li, Zhifeng and Liu, Wei},
  booktitle={International Conference on Learning Representations},
  year={2024}
}

@article{feng2024llmeffichecker,
  title={Llmeffichecker: Understanding and testing efficiency degradation of large language models},
  author={Feng, Xiaoning and Han, Xiaohong and Chen, Simin and Yang, Wei},
  journal={ACM Transactions on Software Engineering and Methodology},
  volume={33},
  number={7},
  pages={1--38},
  year={2024},
  publisher={ACM New York, NY}
}

@article{zou2023universal,
  title={Universal and transferable adversarial attacks on aligned language models},
  author={Zou, Andy and Wang, Zifan and Carlini, Nicholas and Nasr, Milad and Kolter, J Zico and Fredrikson, Matt},
  journal={arXiv preprint arXiv:2307.15043},
  year={2023}
}

@article{frazier2018tutorial,
  title={A tutorial on Bayesian optimization},
  author={Frazier, Peter I},
  journal={arXiv preprint arXiv:1807.02811},
  year={2018}
}

@article{cobbe2021training,
  title={Training verifiers to solve math word problems},
  author={Cobbe, Karl and Kosaraju, Vineet and Bavarian, Mohammad and Chen, Mark and Jun, Heewoo and Kaiser, Lukasz and Plappert, Matthias and Tworek, Jerry and Hilton, Jacob and Nakano, Reiichiro and others},
  journal={arXiv preprint arXiv:2110.14168},
  year={2021}
}

@article{amini2019mathqa,
  title={Mathqa: Towards interpretable math word problem solving with operation-based formalisms},
  author={Amini, Aida and Gabriel, Saadia and Lin, Peter and Koncel-Kedziorski, Rik and Choi, Yejin and Hajishirzi, Hannaneh},
  journal={arXiv preprint arXiv:1905.13319},
  year={2019}
}

@inproceedings{lightman2023let,
  title={Let's verify step by step},
  author={Lightman, Hunter and Kosaraju, Vineet and Burda, Yuri and Edwards, Harrison and Baker, Bowen and Lee, Teddy and Leike, Jan and Schulman, John and Sutskever, Ilya and Cobbe, Karl},
  booktitle={The Twelfth International Conference on Learning Representations},
  year={2023}
}

@article{chen2021evaluating,
  title={Evaluating large language models trained on code},
  author={Chen, Mark},
  journal={arXiv preprint arXiv:2107.03374},
  year={2021}
}

@article{wang2024mmlu,
  title={Mmlu-pro: A more robust and challenging multi-task language understanding benchmark},
  author={Wang, Yubo and Ma, Xueguang and Zhang, Ge and Ni, Yuansheng and Chandra, Abhranil and Guo, Shiguang and Ren, Weiming and Arulraj, Aaran and He, Xuan and Jiang, Ziyan and others},
  journal={Advances in Neural Information Processing Systems},
  volume={37},
  pages={95266--95290},
  year={2024}
}

@article{cuadron2025danger,
  title={The danger of overthinking: Examining the reasoning-action dilemma in agentic tasks},
  author={Cuadron, Alejandro and Li, Dacheng and Ma, Wenjie and Wang, Xingyao and Wang, Yichuan and Zhuang, Siyuan and Liu, Shu and Schroeder, Luis Gaspar and Xia, Tian and Mao, Huanzhi and others},
  journal={arXiv preprint arXiv:2502.08235},
  year={2025}
}

@article{chen2024not,
  title={Do not think that much for 2+ 3=? on the overthinking of o1-like llms},
  author={Chen, Xingyu and Xu, Jiahao and Liang, Tian and He, Zhiwei and Pang, Jianhui and Yu, Dian and Song, Linfeng and Liu, Qiuzhi and Zhou, Mengfei and Zhang, Zhuosheng and others},
  journal={arXiv preprint arXiv:2412.21187},
  year={2024}
}

@article{sui2025stop,
  title={Stop overthinking: A survey on efficient reasoning for large language models},
  author={Sui, Yang and Chuang, Yu-Neng and Wang, Guanchu and Zhang, Jiamu and Zhang, Tianyi and Yuan, Jiayi and Liu, Hongyi and Wen, Andrew and Zhong, Shaochen and Zou, Na and others},
  journal={arXiv preprint arXiv:2503.16419},
  year={2025}
}

@article{kumar2025overthink,
  title={Overthink: Slowdown attacks on reasoning llms},
  author={Kumar, Abhinav and Roh, Jaechul and Naseh, Ali and Karpinska, Marzena and Iyyer, Mohit and Houmansadr, Amir and Bagdasarian, Eugene},
  journal={arXiv preprint arXiv:2502.02542},
  year={2025}
}

@article{rajeev2025cats,
  title={Cats confuse reasoning LLM: Query agnostic adversarial triggers for reasoning models},
  author={Rajeev, Meghana and Ramamurthy, Rajkumar and Trivedi, Prapti and Yadav, Vikas and Bamgbose, Oluwanifemi and Madhusudan, Sathwik Tejaswi and Zou, James and Rajani, Nazneen},
  journal={arXiv preprint arXiv:2503.01781},
  year={2025}
}

@article{agarwal2025gpt,
  title={gpt-oss-120b \& gpt-oss-20b model card},
  author={Agarwal, Sandhini and Ahmad, Lama and Ai, Jason and Altman, Sam and Applebaum, Andy and Arbus, Edwin and Arora, Rahul K and Bai, Yu and Baker, Bowen and Bao, Haiming and others},
  journal={arXiv preprint arXiv:2508.10925},
  year={2025}
}

@misc{llama_companies,
  author = {Meta},
  title = {How Companies Are Using Meta Llama},
  year = {2024},
  url = {https://about.fb.com/news/2024/05/how-companies-are-using-meta-llama/},
}

@inproceedings{rasmussen2003gaussian,
  title={Gaussian processes in machine learning},
  author={Rasmussen, Carl Edward},
  booktitle={Summer school on machine learning},
  pages={63--71},
  year={2003}
}

@inproceedings{movckus1975bayesian,
  title={On Bayesian methods for seeking the extremum},
  author={Mo{\v{c}}kus, Jonas},
  booktitle={Optimization Techniques IFIP Technical Conference},
  pages={400--404},
  year={1975}
}

@article{jones2001taxonomy,
  title={A taxonomy of global optimization methods based on response surfaces},
  author={Jones, Donald R},
  journal={Journal of global optimization},
  volume={21},
  pages={345--383},
  year={2001},
  publisher={Springer}
}

@inproceedings{srinivas2009gaussian,
    author = {Niranjan Srinivas and Andreas Krause and Sham M. Kakade and Matthias W. Seeger},
    bibsource = {dblp computer science bibliography, https://dblp.org},
    booktitle = {International Conference on Machine Learning},
    pages = {1015--1022},
    title = {Gaussian Process Optimization in the Bandit Setting: No Regret and Experimental Design},
    year = {2010}
}

@article{dong2024engorgio,
  title={An engorgio prompt makes large language model babble on},
  author={Dong, Jianshuo and Zhang, Ziyuan and Zhang, Qingjie and Zhang, Tianwei and Wang, Hao and Li, Hewu and Li, Qi and Zhang, Chao and Xu, Ke and Qiu, Han},
  journal={arXiv preprint arXiv:2412.19394},
  year={2024}
}

@article{cheng2024compressed,
  title={Compressed chain of thought: Efficient reasoning through dense representations},
  author={Cheng, Jeffrey and Van Durme, Benjamin},
  journal={arXiv preprint arXiv:2412.13171},
  year={2024}
}

@article{xu2025chain,
  title={Chain of draft: Thinking faster by writing less},
  author={Xu, Silei and Xie, Wenhao and Zhao, Lingxiao and He, Pengcheng},
  journal={arXiv preprint arXiv:2502.18600},
  year={2025}
}

@article{zhang2025pd,
  title={PD3F: A pluggable and dynamic dos-defense framework against resource consumption attacks targeting large language models},
  author={Zhang, Yuanhe and Wang, Xinyue and Gao, Haoran and Zhou, Zhenhong and Meng, Fanyu and Zhang, Yuyao and Su, Sen},
  journal={arXiv preprint arXiv:2505.18680},
  year={2025}
}

@inproceedings{kingma2014adam,
  title={Adam: A method for stochastic optimization},
  author={Kingma, Diederik P},
  booktitle={International Conference on Learning Representations},
  year={2015},
}

@article{jia2024improved,
  title={Improved techniques for optimization-based jailbreaking on large language models},
  author={Jia, Xiaojun and Pang, Tianyu and Du, Chao and Huang, Yihao and Gu, Jindong and Liu, Yang and Cao, Xiaochun and Lin, Min},
  journal={arXiv preprint arXiv:2405.21018},
  year={2024}
}

@inproceedings{kwon2023efficient,
  title={Efficient memory management for large language model serving with pagedattention},
  author={Kwon, Woosuk and Li, Zhuohan and Zhuang, Siyuan and Sheng, Ying and Zheng, Lianmin and Yu, Cody Hao and Gonzalez, Joseph and Zhang, Hao and Stoica, Ion},
  booktitle={Proceedings of the 29th symposium on operating systems principles},
  pages={611--626},
  year={2023}
}

@article{nguyen2024turning,
  title={Turning up the heat: Min-p sampling for creative and coherent llm outputs},
  author={Nguyen, Minh Nhat and Baker, Andrew and Neo, Clement and Roush, Allen and Kirsch, Andreas and Shwartz-Ziv, Ravid},
  journal={arXiv preprint arXiv:2407.01082},
  year={2024}
}

@inproceedings{fan2018hierarchical,
  title={Hierarchical neural story generation},
  author={Fan, Angela and Lewis, Mike and Dauphin, Yann},
  booktitle={Proceedings of the 56th Annual Meeting of the Association for Computational Linguistics (Volume 1: Long Papers)},
  pages={889--898},
  year={2018}
}

@inproceedings{holtzmancurious,
  title={The Curious Case of Neural Text Degeneration},
  author={Holtzman, Ari and Buys, Jan and Du, Li and Forbes, Maxwell and Choi, Yejin},
  booktitle={International Conference on Learning Representations},
  year={2020}
}

@article{meister2022typical,
  title={Typical decoding for natural language generation},
  author={Meister, Clara and Pimentel, Tiago and Wiher, Gian and Cotterell, Ryan},
  journal={arXiv preprint arXiv:2202.00666},
  volume={4},
  year={2022}
}

@article{dettmers2023qlora,
  title={Qlora: Efficient finetuning of quantized llms},
  author={Dettmers, Tim and Pagnoni, Artidoro and Holtzman, Ari and Zettlemoyer, Luke},
  journal={Advances in neural information processing systems},
  volume={36},
  pages={10088--10115},
  year={2023}
}

@inproceedings{merrill2026terminal,
  title={Terminal-bench: Benchmarking agents on hard, realistic tasks in command line interfaces},
  author={Merrill, Mike and Shaw, Alexander and Carlini, Nicholas and Li, Boxuan and Raj, Harsh and Bercovich, Ivan and Shi, Lin and Shin, Jeong and Walshe, Thomas and Buchanan, E Kelly and others},
  booktitle={International Conference on Learning Representations},
  volume={2026},
  pages={40903--40986},
  year={2026}
}

@article{phan2025lastexam,
      title = {A benchmark of expert-level academic questions to assess {AI} capabilities},
      author = {{Center for AI Safety} and {Scale AI} and {HLE Contributors Consortium}},
      journal = {Nature},
      volume = {649},
      pages = {1139--1146},
      year = {2026},
      doi = {10.1038/s41586-025-09962-4},
      eprint = {2501.14249},
      archivePrefix = {arXiv},
      primaryClass = {cs.LG},
      url = {https://arxiv.org/abs/2501.14249}
}

@inproceedings{patwardhan2026gdpval,
  title={Gdpval: Evaluating ai model performance on real-world economically valuable tasks},
  author={Patwardhan, Tejal and Dias, Rachel and Proehl, Elizabeth and Kim, Grace and Wang, Michele and Watkins, Olivia and Fishman, Simon and Aljubeh, Marwan and Thacker, Phoebe and Fauconnet, Laurance and others},
  booktitle={International Conference on Learning Representations},
  volume={2026},
  pages={24005--24040},
  year={2026}
}

@article{xu2022learning,
  title={Learning to break the loop: Analyzing and mitigating repetitions for neural text generation},
  author={Xu, Jin and Liu, Xiaojiang and Yan, Jianhao and Cai, Deng and Li, Huayang and Li, Jian},
  journal={Advances in Neural Information Processing Systems},
  volume={35},
  pages={3082--3095},
  year={2022}
}

@article{szegedy2013intriguing,
  title={Intriguing properties of neural networks},
  author={Szegedy, Christian and Zaremba, Wojciech and Sutskever, Ilya and Bruna, Joan and Erhan, Dumitru and Goodfellow, Ian and Fergus, Rob},
  journal={arXiv preprint arXiv:1312.6199},
  year={2013}
}

@inproceedings{carlini2017towards,
  title={Towards evaluating the robustness of neural networks},
  author={Carlini, Nicholas and Wagner, David},
  booktitle={2017 IEEE symposium on security and privacy (SP)},
  pages={39--57},
  year={2017},
  organization={IEEE}
}

@inproceedings{dong2018boosting,
  title={Boosting adversarial attacks with momentum},
  author={Dong, Yinpeng and Liao, Fangzhou and Pang, Tianyu and Su, Hang and Zhu, Jun and Hu, Xiaolin and Li, Jianguo},
  booktitle={Proceedings of the IEEE/CVF Conference on Computer Vision and Pattern Recognition},
  pages={9185--9193},
  year={2018}
}

@inproceedings{miao2022isometric,
  title={Isometric 3D adversarial examples in the physical world},
  author={Miao, Yibo and Dong, Yinpeng and Zhu, Jun and Gao, Xiao-Shan},
  booktitle={Advances in Neural Information Processing Systems},
  volume={35},
  pages={19716--19731},
  year={2022}
}

@inproceedings{cheng2024efficient,
  title={Efficient black-box adversarial attacks via Bayesian optimization guided by a function prior},
  author={Cheng, Shuyu and Miao, Yibo and Dong, Yinpeng and Yang, Xiao and Gao, Xiao-Shan and Zhu, Jun},
  booktitle={International Conference on Machine Learning},
  pages={8163--8183},
  year={2024}
}

@article{miao2024improving,
  title={Improving Robustness of 3D Point Cloud Recognition from a Fourier Perspective},
  author={Miao, Yibo and Dong, Yinpeng and Zhang, Jinlai and Yu, Lijia and Yang, Xiao and Gao, Xiao-Shan},
  journal={Advances in Neural Information Processing Systems},
  volume={37},
  pages={68183--68210},
  year={2024}
}

@inproceedings{zhu2024toward,
  title={Toward availability attacks in 3D point clouds},
  author={Zhu, Yifan and Miao, Yibo and Dong, Yinpeng and Gao, Xiao-Shan},
  booktitle={International Conference on Machine Learning},
  pages={62510--62530},
  year={2024}
}

@inproceedings{yu2024generalization,
  title={Generalization bound and new algorithm for clean-label backdoor attack},
  author={Yu, Lijia and Liu, Shuang and Miao, Yibo and Gao, Xiao-Shan and Zhang, Lijun},
  booktitle={International Conference on Machine Learning},
  pages={57559--57596},
  year={2024}
}

@inproceedings{miao2024t2vsafetybench,
  title={T2Vsafetybench: Evaluating the safety of text-to-video generative models},
  author={Miao, Yibo and Zhu, Yifan and Yu, Lijia and Zhu, Jun and Gao, Xiao-Shan and Dong, Yinpeng},
  booktitle={Advances in Neural Information Processing Systems},
  volume={37},
  pages={63858--63872},
  year={2024}
}

@article{cao2026red,
  title={Red-teaming text-to-image systems by rule-based preference modeling},
  author={Cao, Yichuan and Miao, Yibo and Gao, Xiao-Shan and Dong, Yinpeng},
  journal={Advances in Neural Information Processing Systems},
  volume={38},
  pages={150703--150734},
  year={2026}
}

@inproceedings{zhu2026unlearnable,
  title={Why do unlearnable examples work: A novel perspective of mutual information},
  author={Zhu, Yifan and Miao, Yibo and Dong, Yinpeng and GAO, XIAOSHAN},
  booktitle={International Conference on Learning Representations},
  volume={2026},
  pages={107849--107880},
  year={2026}
}

@article{yu2024generalizablity,
  title={Generalizablity of memorization neural network},
  author={Yu, Lijia and Gao, Xiao-Shan and Zhang, Lijun and Miao, Yibo},
  journal={Advances in Neural Information Processing Systems},
  volume={37},
  pages={113311--113359},
  year={2024}
}
\bibliographystyle{plainnat}


\newpage
\appendix

\section{Impact Statement}

A potential negative societal impact of our work is that malicious adversaries could misuse the proposed resource consumption attacks (Chatter Attack and Chatter-BPO  Attack) against deployed large language model (LLM) services. By inducing unusually long generations, such attacks can increase inference latency and operational costs, degrade quality of service, and potentially contribute to denial-of-service-like outages that harm legitimate users. These risks are particularly relevant for publicly accessible or pay-per-use APIs, and they may also increase energy consumption and carbon emissions. To mitigate these risks, our paper evaluates representative mitigation/defense strategies and highlights practical monitoring signals (e.g., request concurrency, time-to-first-token, and token throughput) that can help operators detect and respond to abnormal long-generation behaviors. We encourage service providers to combine these insights with deployment safeguards such as output-length limits, per-user rate limiting, anomaly detection, and server-side termination policies. To further reduce misuse, we will avoid releasing directly weaponizable attack prompts or code that enables unattended large-scale abuse, and we will share relevant findings with affected organizations to support more secure and reliable LLM deployments.

\section{Related Work} 

\textbf{Reasoning Large Language Model.}
Reasoning LLMs, such as OpenAI o1/o3~\citep{jaech2024openai}, DeepSeek-R1 \citep{guo2025deepseek}, and QWQ~\citep{team2025qwq}, enhance the handling of multi-step reasoning tasks, including mathematics \citep{ahn2024large}, programming~\citep{yang2023coupling}, and logical reasoning~\citep{liu2025logical}, by integrating explicit multi-step reasoning chains~\citep{wei2022chain} into LLMs. However, as the reasoning capability of these models increases, the computational resource consumption during the inference stage becomes a significant concern. Previous studies indicate that when the query volume exceeds several hundred million, the inference cost surpasses the training cost, making large-scale deployment a major energy consumer~\citep{patterson2022carbon}. Furthermore, these models are prone to experiencing the "overthinking" phenomenon~\citep{cuadron2025danger, chen2024not, sui2025stop}, where unnecessary lengthy reasoning steps are generated during inference. \citet{chen2024not} study this phenomenon and show that LRMs often expend excessive computational resources on simple problems. In this study, we adopt the role of a malicious attacker, exploring how to exploit this characteristic to induce lengthy outputs, thereby increasing response latency and inference costs.

\textbf{Resource Consumption Attack.}
Deep learning safety has been extensively studied~\cite{szegedy2013intriguing, carlini2017towards, dong2018boosting, miao2022isometric, cheng2024efficient, miao2024improving, zhu2024toward, yu2024generalization, miao2024t2vsafetybench, cao2026red, zhu2026unlearnable, yu2024generalizablity}.
Building on this,
resource consumption attacks are first introduced by~\citet{shumailov2021sponge}, customized for encoder-decoder architectures, by maximizing the activation norm to increase computational costs, thereby maximizing energy consumption and latency during model inference. NICGSlowDown~\citep{chen2022nicgslowdown} manipulates the logits of output tokens, increasing the number of decoder calls to reduce inference efficiency.
However, Sponge Samples~\citep{shumailov2021sponge} are specifically designed for translation tasks, while NICGSlowDown targets image captioning tasks for Recurrent Neural Networks (RNNs) or Long Short-Term Memory networks (LSTMs). These methods are not directly applicable to modern large language models (LLMs) that adopt autoregressive generation schemes.
OverThink Attack~\citep{kumar2025overthink} and CatAttack~\citep{rajeev2025cats} introduce semantic manipulation, such as injecting inherently open-ended problem prompts or complex mathematical problems, causing the model to allocate additional reasoning resources to solve the baited prompts. While occasionally effective, these techniques generally rely on handcrafted templates, lacking robustness and generalization capabilities, and demonstrate effectiveness only on specific models.
\citet{geiping2024coercing} propose an adversarial prompt attack, setting the attack target as repeated vocabulary, coercing LLMs to repeat specific content, thereby increasing inference latency by extending the model's output length. Similarly, methods like Engorgio~\citep{dong2025engorgio, si2025excessive, yan2025bithydra, gao2024inducing, feng2024llmeffichecker} minimize the logits of the end-of-sequence (EOS) token relative to each output token, suppressing the EOS token's appearance and generating excessively long text outputs, imposing a computational burden on the model.
Since LLMs are autoregressive and sampling-based, the search space of these methods grows exponentially with the number of tokens due to the inherent randomness, making it difficult to optimize effectively.

\section{Detailed Experiment Settings}
\label{App:experiment_settings}

\textbf{Datasets.} We evaluate our attacks on five benchmarks covering math reasoning, general-knowledge question answering, and code generation: the grade school level GSM8K~\citep{cobbe2021training}, multiple choices problems MathQA~\citep{amini2019mathqa}, high school level mathematical datasets MATH-500~\citep{lightman2023let}, challenging question-answering benchmark MMLU-Pro~\citep{wang2024mmlu}, and code generation benchmark HumanEval~\citep{chen2021evaluating}.
For GSM8K, MathQA, MATH-500, and HumanEval, we randomly sample 100 instances per dataset. For MMLU-Pro, we select 10 questions from each of the 12 non-math categories (for example, chemistry, biology).

\textbf{Models.} We assess our attack using three state-of-the-art reasoning models: DeepSeek-R1-Distilled-Qwen-7b (R1-7b)~\citep{guo2025deepseek}, QwQ-32b~\citep{team2025qwq}, and gpt-oss-20b~\citep{agarwal2025gpt}. For model generation, we generate up to 16384 tokens with temperature $0.6$, top-$p$ $0.95$, and repetition penalty $1.0$, which is a commonly used configuration for evaluating the capability of model reasoning. To simulate a realistic attack scenario, we additionally deploy a local vLLM inference server to mimic a commercial LLM service provider. We deploy the model using vLLM~\citep{kwon2023efficient} on one NVIDIA A800 GPU and issue queries with the same decoding parameters.

\textbf{Baselines.} We compare our attack against the state-of-the-art resource consumption attacks that induce excessively long outputs and increase inference latency.
Specifically, we include multiple black-box baselines: chain-of-thought (CoT) prompting~\citep{wei2022chain}, CatAttack~\citep{rajeev2025cats}, and four variants of OverThink~\citep{kumar2025overthink}, including Context-Aware (CAW), Context-Agnostic (CAG), ICL-Genetic-Aware (ICLGAW), and ICL-Genetic-Agnostic (ICLGAG). For CoT prompting, we instruct the model to reason step-by-step using the prompt template illustrated in Table~\ref{app_tab:cot_template}. For two ICL-Genetic variants of OverThink, we use the GPT-5-mini for adversarial context generation and perform 10 rounds optimization. For white-box baselines, we compare with the Denial-of-Service (DoS) prompt attack~\citep{geiping2024coercing}, Engorgio Prompt~\citep{dong2024engorgio}, and Excessive Reasoning Attack (ERA)~\citep{si2025excessive}. For DoS attack, we perform GCG algorithm with the repetition of the string ``Hello There'' 24 times as the objectives, and optimize with 64 tokens suffix. For Engorgio Prompt, we adapt a pre-set maximum length 2048 for all reasoning models. For Excessive Reasoning Attack, we optimize the 10 token suffix with 1000 steps GCG. All experiments are performed exactly according to their experimental configuration respectively.

\begin{table*}[ht!]
\centering
\caption{Prompt template for CoT prompting.}
\label{app_tab:cot_template}
\resizebox{0.6\textwidth}{!}{
\begin{tabular}{l}
\Xhline{1.25pt}
\begin{tabular}[c]{@{}l@{}}
\textbf{Base prompting:}\\
\textless ORIGINAL QUESTION \textgreater\\
Please put your answer in 	
\textbackslash boxed\{\}. \\
\Xhline{1.25pt}
\textbf{CoT prompting:}\\
\textless ORIGINAL QUESTION \textgreater\\
Please \textcolor{red}{reason step by step} and put your answer in 	
\textbackslash boxed\{\}. \\

\end{tabular} \\ 
\Xhline{1.25pt}
\end{tabular}%
}
\end{table*}

\begin{figure*}[t!]
    \centering
    \includegraphics[width=\textwidth]{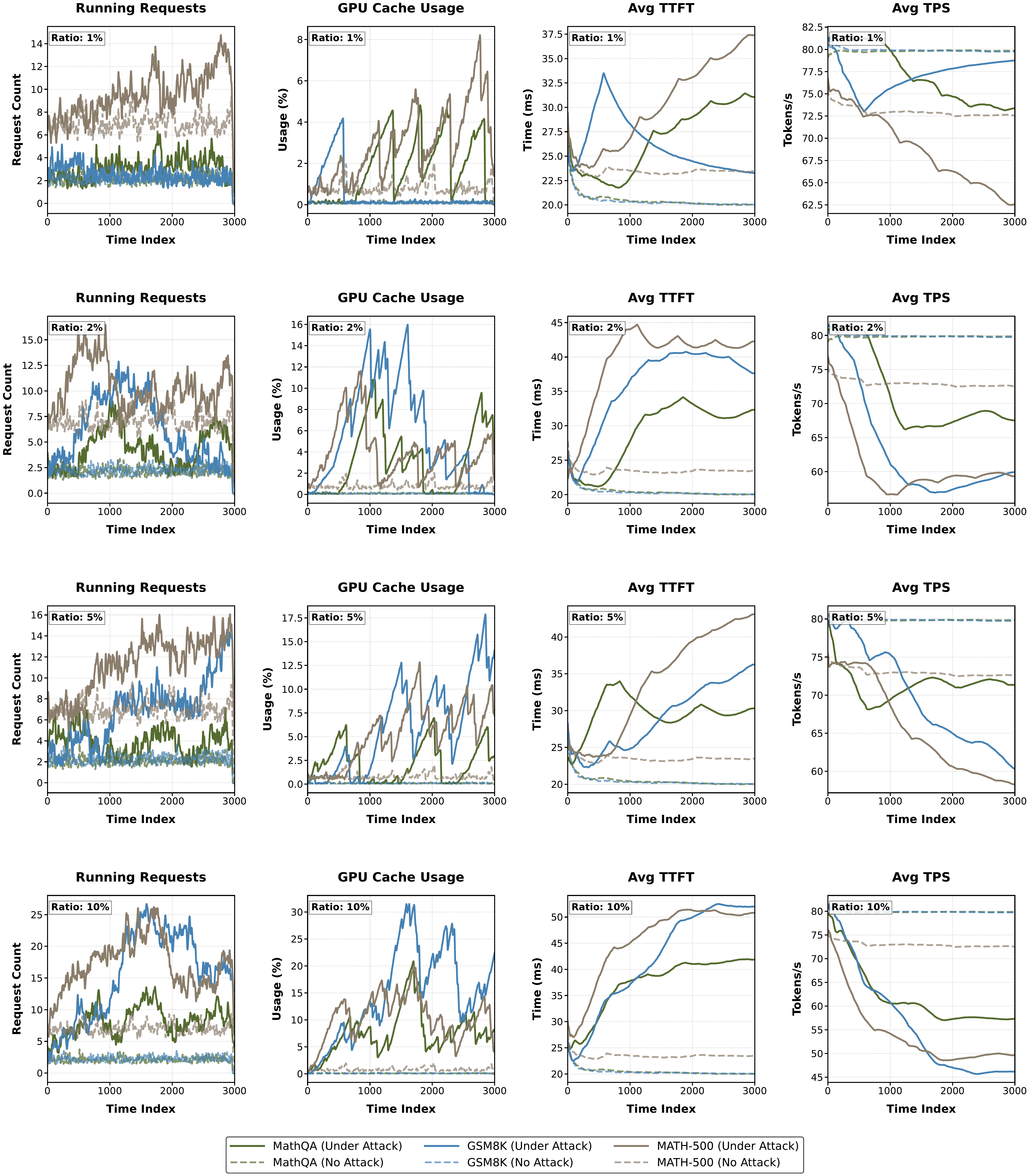}
    \caption{
        Server-side runtime metrics of the deployed LLM service under purely benign traffic and mixed traffic containing 1\% to 10\% Chatter Attack queries. The service is evaluated at a constant rate of 20 requests per minute.
    }
    \label{fig:app_server_1}
\end{figure*}

Additionally, we also develop three extra baselines to provide a more direct and transparent comparison of optimization strategies. For the Black-box Search baseline, we prompt GPT-5-mini to iteratively refine the current prefix to maximize the output length. The Gradient-free Attack performs a 3,000-step random search to construct the prefix, modifying one token per step in the direction that increases the output length. Heuristic Prompt is a heuristic prompt-based method inducing longer output, with the prompt template shown in Table~\ref{tab:induction_template}. 

\begin{table*}[ht!]
\centering
\caption{Prompt template for Heuristic Prompt.}
\label{tab:induction_template}
\resizebox{\textwidth}{!}{
\begin{tabular}{l}
\Xhline{1.25pt}
\begin{tabular}[c]{@{}l@{}}
Please answer the following QUESTION:\\
QUESTION: \textless ORIGINAL QUESTION \textgreater\\\\
You should follow these guidelines strictly:\\
1. Multi-Dimensional Deconstruction: Thoroughly decompose the problem into its constituent parts, \\examining the core issue and all peripheral factors from a panoramic perspective.\\
\\
2. Explicit Logical Chain: Externalize every micro-step of your reasoning process. \\Avoid logical leaps and ensure that the transition between every premise and conclusion is fully articulated with high-density detail.\\
\\
3. Recursive Layering of Depth: After reaching an initial conclusion, drill down into deeper layers of "Why" and "How."\\ Analyze the implications within various theoretical frameworks, historical contexts, or extreme edge cases.\\
\\
4. Exhaustive Scenario Simulation: Explore not only the most probable outcome but also all alternative interpretations and counter-narratives. \\Provide a detailed walkthrough of the mechanics, causes, and consequences of each possibility.\\
\\
5. Incremental Enrichment: For every concept or term introduced, provide a comprehensive background explanation \\or contextual definition before proceeding with the main argument to ensure total clarity.\\
\\
6. Iterative Expansion: Upon completing a section, revisit it to identify areas where information density can be increased. \\Force-multiply the output by adding illustrative examples, data-driven analogies, or comparative analysis until maximum granularity is achieved.
\\\\
Please reason step by step and put your answer in 	
\textbackslash boxed\{\}. \\
\end{tabular} \\ 
\Xhline{1.25pt}
\end{tabular}%
}
\end{table*}

\textbf{Defenses.} To more comprehensively evaluate the performance of our attack under defenses, we select several representative methods for reasoning compression and resource-consumption mitigation, including the prompt-based CCoT~\citep{cheng2024compressed}, CoD~\citep{xu2025chain}, and decoding-based PD$^3$F~\citep{zhang2025pd}. CCoT and CoD use an explict prompt to instruct model keep concise for the output, while PD$^3$F develop a complex systems to monitor the output budget of LLM. 

Besides, we also implement and evaluate on multiple different production constraints, traditional infrastructure defenses, and post-generation defenses in a simulated server environment. All evaluations share the same experimental setup: queries are injected continuously at intervals of 2--4s over a total duration of 3000s, with malicious attack requests accounting for 10\% of the traffic and the remaining 90\% consisting of lightweight background traffic from 200 independent benign accounts. The attacker is assigned a conservative resource pool consisting of 5 IPs and 20 accounts, with each IP hosting 4 accounts.
For the production constraints, strict output-token limits cap every request's max\_tokens at 8,192 at admission. Per-user rate limits enforce token-bucket-based limits with RPM $=20$ and TPM $=10$k. Dynamic batching dynamically varies vLLM's concurrency limit between 256 and 512. Request prioritization maps account reputation, derived from registration age, historical output length, and violation records, to priority levels in vLLM's native scheduler. Server-side early termination policies use a lightweight online trigger based on cumulative self-doubt markers or generated length to abort requests in real time and release resources.
For traditional infrastructure defenses, spend caps are set to a \$5 trial budget per account, equivalent to a quota of $\sim$100k output tokens based on simulated unit pricing. IP/account blocking is triggered by 3 consecutive outputs exceeding 5000 tokens, resulting in a 600s blocking. These threshold settings are highly realistic, as they both ensure legitimate usage by normal users and reflect actual defense baselines.
Anomaly detection covers both the input and output sides. On the input side, we compute prompt perplexity using Qwen2.5-7B-Instruct to filter high-PPL inputs. On the output side, loop detection follows classic degeneration criteria~\citep{holtzmancurious, xu2022learning}, flagging an output when its repeated 4-gram ratio exceeds 0.5 or its gzip compression ratio falls below 0.10. Excessive-reasoning monitoring is implemented using an LLM-as-a-judge (Qwen2.5-7B-Instruct) that classifies each completed CoT as overthinking or normal based on the overthinking criteria in~\citep{chen2024not}. In both output-side cases, the enforcement policy is aligned with our strictest account-level strategy: an account is banned for 600s after accumulating 3 flagged outputs.

\textbf{Metrics.} To evaluate the effectiveness of our attack, we report the widely-used metrics including average output token length (ATL), average inference time consumption (ATC), and accuracy (ACC), with the 95\% confidence intervals of ATL and ATC.
Additionally, to mimic realistic attacks against API services and assess the impact of Chatter Attack on server load and responsiveness, we also monitor the server-side runtime indicators during inference including the counts of running requests (RR), GPU cache usage (GCU), response time to first token (TTFT), and the generated token counts per second (TPS). RR and GCU represent the state of server load, while TTFT and TPS directly reflect the velocity of response and generation for the service.

\begin{table*}[t]
\centering
\caption{
Quantitative results of baselines and our attacks on question-answer and coding benchmarks on R1-7b. Chatter Attack and Chatter-BPO Attack could generalize to these extensive reasoning settings.
}
\label{tab:other_reasoning}
\resizebox{0.8\textwidth}{!}{%
\begin{tabular}{l|ccc|ccc}
\toprule
                            & \multicolumn{3}{c|}{MMLU-Pro}                     & \multicolumn{3}{c}{HumanEval}                      \\
                            & ATL                   & ATC                & ACC  & ATL                    & ATC                & ACC  \\ \midrule
Base                        & 1495.0 $\pm$ 492.0          & 15.6 $\pm$ 3.7           & 28.7 & 1554.7 $\pm$ 138.1           & 17.9 $\pm$ 1.6           & 98.3 \\
CoT                         & 1550.6 $\pm$ 429.2          & 17.0 $\pm$ 3.8           & 26.9 & 1691.2 $\pm$ 167.1           & 19.4 $\pm$ 1.9           & 98.8 \\
DoS                         & 1220.2 $\pm$ 401.7          & 13.9 $\pm$ 3.6           & 16.3 & 1352.2 $\pm$ 92.9            & 16.4 $\pm$ 1.1           & 97.4 \\
Engorgio   Prompt           & 1760.8 $\pm$ 419.2          & 19.7 $\pm$ 3.8           & 25.5 & 1287.9 $\pm$ 181.1           & 14.9 $\pm$ 2.1           & 97.9 \\
CAW                         & 861.5 $\pm$ 119.2           & 9.9 $\pm$ 1.4            & 15.1 & 1453.8 $\pm$ 123.5           & 16.7 $\pm$ 1.4           & 99.3 \\
CAG                         & 767.6 $\pm$ 94.9            & 9.3 $\pm$ 1.1            & 7.8  & 1430.9 $\pm$ 112.9           & 17.3 $\pm$ 1.4           & 99.5 \\
ICLGAW                      & 924.5 $\pm$ 157.4           & 11.0 $\pm$ 1.9           & 17.5 & 3297.1 $\pm$ 906.2           & 34.3 $\pm$ 8.2           & 85.4 \\
ICLGAG                      & 774.4 $\pm$ 86.3            & 9.2 $\pm$ 1.0            & 15.5 & 1327.0 $\pm$ 101.0           & 15.7 $\pm$ 1.2           & 99.8 \\
CatAttack                   & 1306.2 $\pm$ 184.2          & 15.0 $\pm$ 2.1           & 16.1 & 1646.7 $\pm$ 145.1           & 18.9 $\pm$ 1.7           & 98.0 \\
ERA                         & 757.0 $\pm$ 54.3            & 8.9 $\pm$ 0.6            & 5.9  & 3195.4 $\pm$ 993.9           & 29.8 $\pm$ 7.0           & 88.3 \\
Gradient-free   Attack      & 1170.1 $\pm$ 151.9          & 13.3 $\pm$ 1.7           & 15.4 & 1463.3 $\pm$ 98.3            & 17.0 $\pm$ 1.2           & 99.1 \\
Black-box   Search          & 1309.7 $\pm$ 209.5          & 13.9 $\pm$ 2.1           & 19.7 & 2538.1 $\pm$ 120.5           & 29.8 $\pm$ 1.4           & 97.6 \\
Heuristic   Prompt          & 1333.4 $\pm$ 234.9          & 14.4 $\pm$ 2.5           & 14.0 & 2221.5 $\pm$ 184.4           & 23.5 $\pm$ 2.0           & 98.0 \\
Chatter   Attack (ours)     & \textbf{6638.3 $\pm$ 962.4} & \textbf{70.2 $\pm$ 11.6} & 5.3  & \textbf{5788.4 $\pm$ 1400.3} & \ul{63.4 $\pm$ 17.0}    & 43.3 \\
Chatter-BPO   Attack (ours) & \ul{4251.1 $\pm$ 1159.5}   & \ul{49.3 $\pm$ 13.7}    & 4.4  & \ul{5516.2 $\pm$ 1285.6}    & \textbf{64.2 $\pm$ 15.2} & 44.5 \\ \bottomrule
\end{tabular}
}
\end{table*}

\textbf{Implementation Details.} For the implementation of Chatter Attack, we use Greedy Coordinate Gradient (GCG)~\citep{zou2023universal} to optimize an adversarial prefix $a$ with a length of 20 tokens for 3000 steps, and use a sample budget of 64. We set the hyperparameters of the overall objective function $\mathcal{L}$ to $(\alpha, \beta, \lambda) = (1, 10, 1)$. Furthermore, we employ the vocabulary presented in Table~\ref{App:word_list} to identify the critical reasoning tokens for inference attention loss. For the implementation of Chatter-BPO Attack, we use the gpytorch to implement the kernel. We utilize the Adam optimizer~\citep{kingma2014adam} with a learning rate of $0.1$ to optimize the hyperparameters of the Gaussian process. We set $\gamma = 2$. We maximize the acquisition function using 100 steps GCG, and then apply a dynamic random-search schedule with gradually increasing intensity from no search $N=0$ to $N=16$ steps iterations, to enable more efficient exploration.

\begin{table*}[ht!]
\centering
\caption{Critical reasoning words vocabulary for inference attention loss.}
\label{App:word_list}
\resizebox{0.5\textwidth}{!}{
\begin{tabular}{l}
\Xhline{1.25pt}
\begin{tabular}[c]{@{}l@{}}

Wait\quad Alternatively\quad  But\quad   However\quad  Hmm\quad \\ 
wait\quad  alternatively\quad  but\quad  however\quad  hmm 

\end{tabular} \\ 
\Xhline{1.25pt}
\end{tabular}%
}
\end{table*}

\begin{figure*}[t]
    \centering
    \includegraphics[width=\textwidth]{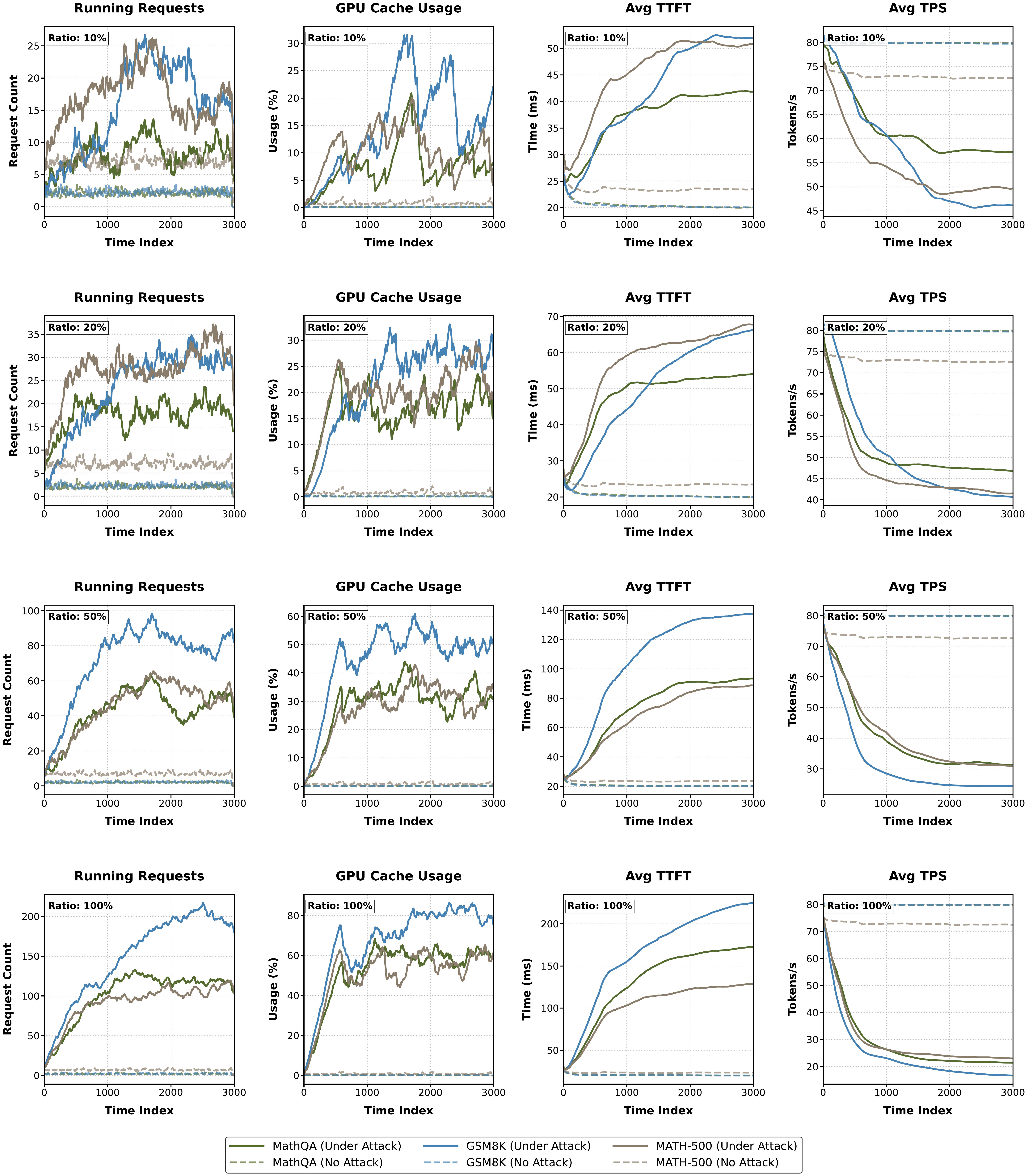}
    \caption{
        Server-side runtime metrics of the deployed LLM service under purely benign traffic and mixed traffic containing 10\% to 100\% Chatter Attack queries. The service is evaluated at a constant rate of 20 requests per minute.
    }
    \label{fig:app_server_2}
\end{figure*}

\begin{table*}[t]
\centering
\caption{
Ablation studies of the loss function for Chatter Attack.
}
\label{tab:loss_ablation}
\resizebox{0.7\textwidth}{!}{%
\begin{tabular}{l|ccc}
\toprule
                                                                            & ATL            & ATC        & ACC  \\ \midrule
Chatter   Attack                                                            & 13345.8 $\pm$ 1067.4 & 105.4 $\pm$ 16.4 & 70.4 \\
Chatter Attack w/o $\mathcal{L}_{\text{IA}}$                                & 9545.6 $\pm$ 1285.1  & 96.6 $\pm$ 15.2  & 78.9 \\
Chatter Attack w/o $\mathcal{L}_{\text{GE}}$                                & 9282.3 $\pm$ 1400.5  & 93.9 $\pm$ 16.7  & 77.2 \\
Chatter Attack w/o $\mathcal{L}_{\text{IA}}$   \& $\mathcal{L}_{\text{GE}}$ & 8088.0 $\pm$ 1237.0  & 89.3 $\pm$ 14.6  & 75.6 \\ \bottomrule
\end{tabular}
}
\end{table*}

\begin{table*}[t]
\centering
\caption{
Sensitivity analyses of the hyperparameters for the loss function of Chatter Attack. The numbers in the brackets represent the hyperparameter combinations $(\alpha, \beta, \lambda)$.
}
\label{tab:loss_sensitive}
\resizebox{0.7\textwidth}{!}{%
\begin{tabular}{l|ccc}
\toprule
                                      & ATL            & ATC        & ACC  \\ \midrule
Chatter   Attack (1, 10, 1) (default) & 13345.8 $\pm$ 1067.4 & 105.4 $\pm$ 16.4 & 70.4 \\
Chatter   Attack (1, 5, 1)            & 7323.2 $\pm$ 1510.4  & 80.3 $\pm$ 18.7  & 48.9 \\
Chatter   Attack (1, 7.5, 1)          & 10123.8 $\pm$ 1325.2 & 101.9 $\pm$ 17.3 & 74.4 \\
Chatter   Attack (1, 12.5, 1)         & 9845.8 $\pm$ 1177.2  & 98.1 $\pm$ 15.9  & 67.8 \\
Chatter   Attack (1, 15, 1)           & 8228.8 $\pm$ 1351.9  & 93.5 $\pm$ 17.6  & 63.3 \\
Chatter   Attack (1, 10, 0.5)         & 8516.7 $\pm$ 1436.7  & 94.3 $\pm$ 18.2  & 64.8 \\
Chatter   Attack (1, 10, 0.75)        & 9105.5 $\pm$ 1431.0  & 97.1 $\pm$ 18.0  & 70.9 \\
Chatter   Attack (1, 10, 1.25)        & 9967.3 $\pm$ 1514.5  & 101.2 $\pm$ 18.7 & 75.6 \\
Chatter   Attack (1, 10, 1.5)         & 8680.5 $\pm$ 1395.8  & 87.8 $\pm$ 17.1  & 72.2 \\ \bottomrule
\end{tabular}
}
\end{table*}

\begin{table*}[b]
\centering
\caption{
Sensitivity analyses of the token length for the attack prefix. The number in the brackets represent the attack prefix length.
}
\label{tab:prefix_length}
\resizebox{0.7\textwidth}{!}{%
\begin{tabular}{l|ccc}
\toprule
 & ATL & ATC & ACC \\ \midrule
Chatter Attack (20) (default) & 13345.8 $\pm$ 1067.4 & 105.4 $\pm$ 16.4 & 70.4 \\
Chatter Attack (10) & 7846.9 $\pm$ 1505.0 & 80.3 $\pm$ 12.7 & 75.6 \\
Chatter Attack (15) & 9555.1 $\pm$ 1474.4 & 91.5 $\pm$ 14.1 & 68.6 \\
Chatter Attack (25) & 13346.9 $\pm$ 1374.3 & 104.6 $\pm$ 19.6 & 78.5 \\
Chatter Attack (30) & 14147.5 $\pm$ 1084.8 & 119.5 $\pm$ 16.2 & 71.3 \\ \bottomrule
\end{tabular}%
}
\end{table*}

\begin{table*}[t]
\centering
\caption{
Ablation studies of the Chatter-BPO Attack on GSM8K on R1-7b.
}
\label{tab:bpo_ablation}
\resizebox{0.7\textwidth}{!}{%
\begin{tabular}{l|ccc}
\toprule
 & ATL & ATC & ACC  \\ 
\midrule
Random Search & 1532.4 $\pm$ 345.2 & 18.3 $\pm$ 5.1 & 81.2 \\
Chatter-BPO w/o Local Search & 5215.8 $\pm$ 890.1 & 45.6 $\pm$ 12.3 & 73.8 \\
Chatter-BPO Attack & 7846.9 $\pm$ 1473.5 & 91.7 $\pm$ 17.4 & 67.6 \\ 
\bottomrule
\end{tabular}%
}
\end{table*}

\section{Additional Experiments}

\begin{table*}[t]
\centering
\caption{
Robustness evaluation of Chatter Attack under various decoding settings. The evaluated configurations included different decoding strategies, temperature choices, and Top-P values. Chatter Attack consistently induces high ATL across diverse decoding settings, exhibiting its strong robustness.
}
\label{tab:decoding}
\resizebox{0.7\textwidth}{!}{%
\begin{tabular}{cccc}
\toprule
\multicolumn{4}{c}{Decoding Strategies} \\ \midrule
\multicolumn{1}{l|}{} & ATL & ATC & ACC \\ \midrule
\multicolumn{1}{l|}{Top-P Decoding (default)} & 13345.8 $\pm$ 1067.4 & 105.4 $\pm$ 16.4 & 70.4 \\
\multicolumn{1}{l|}{Greedy Decoding} & 14796.6 $\pm$ 976.0 & 110.2 $\pm$ 14.9 & 56.7 \\
\multicolumn{1}{l|}{MinP Decoding} & 12313.3 $\pm$ 1200.8 & 98.6 $\pm$ 17.2 & 66.7 \\
\multicolumn{1}{l|}{Top-K Decoding} & 12980.4 $\pm$ 1122.0 & 103.1 $\pm$ 16.8 & 65.6 \\
\multicolumn{1}{l|}{Typical Decoding} & 12195.4 $\pm$ 1273.3 & 96.9 $\pm$ 18.5 & 64.4 \\ \midrule
\multicolumn{4}{c}{Temperature} \\ \midrule
\multicolumn{1}{l|}{} & ATL & ATC & ACC \\ \midrule
\multicolumn{1}{l|}{1.0} & 8356.0 $\pm$ 869.6 & 80.5 $\pm$ 12.4 & 74.4 \\
\multicolumn{1}{l|}{0.8} & 10654.1 $\pm$ 1222.5 & 96.2 $\pm$ 18.3 & 66.7 \\
\multicolumn{1}{l|}{0.6 (default)} & 13345.8 $\pm$ 1067.4 & 105.4 $\pm$ 16.4 & 70.4 \\
\multicolumn{1}{l|}{0.4} & 13942.4 $\pm$ 1095.0 & 105.3 $\pm$ 17.3 & 64.4 \\
\multicolumn{1}{l|}{0.2} & 14983.7 $\pm$ 833.3 & 112.2 $\pm$ 14.3 & 62.2 \\ \midrule
\multicolumn{4}{c}{Top-P Decoding} \\ \midrule
\multicolumn{1}{c|}{} & ATL & ATC & ACC \\ \midrule
\multicolumn{1}{l|}{1.0} & 11131.6 $\pm$ 1246.7 & 92.3 $\pm$ 18.0 & 65.6 \\
\multicolumn{1}{l|}{0.99} & 11783.9 $\pm$ 1289.3 & 94.8 $\pm$ 18.5 & 68.9 \\
\multicolumn{1}{l|}{0.95 (default)} & 13345.8 $\pm$ 1067.4 & 105.4 $\pm$ 16.4 & 70.4 \\
\multicolumn{1}{l|}{0.7} & 14884.9 $\pm$ 883.8 & 110.9 $\pm$ 14.7 & 68.9 \\
\multicolumn{1}{l|}{0.5} & 14825.2 $\pm$ 910.9 & 110.6 $\pm$ 16.2 & 67.8 \\ \bottomrule
\end{tabular}%
}
\end{table*}

\subsection{Ablation Studies of Chatter Attack's Objective Function}
\label{app:ablation}

We adopt the following objective function for Chatter Attack:
\begin{equation}
\mathcal{L} = \alpha \mathcal{L}_{\text{ID}} + \beta \mathcal{L}_{\text{IA}} + \lambda \mathcal{L}_{\text{GE}}
\end{equation}
To better understand the contribution of each term, we perform ablation studies on the three components of the Chatter Attack objective. As shown in Table~\ref{tab:loss_ablation}, combining $\mathcal{L}_{\text{ID}}$ with either $\mathcal{L}_{\text{IA}}$ or $\mathcal{L}_{\text{GE}}$ consistently outperforms using $\mathcal{L}_{\text{ID}}$ alone. Moreover, the full objective that includes all three losses achieves the best performance, suggesting that these terms are complementary.

\subsection{Sensitivity Analyses of the Hyperparameters for the Loss Function}
\label{app:sensitive}

We conduct a detailed sensitivity analysis of the hyperparameters in the Chatter Attack objective function. Specifically, we fix $\alpha=1$ and vary $\beta$ and $\lambda$, optimizing the attack on GSM8K with R1-7b. As shown in Table~\ref{tab:loss_sensitive}, the default configuration achieves the best performance; meanwhile, across a wide range of hyperparameter settings, our attack consistently attains stronger optimization results than the baselines (Table~\ref{tab:main}), indicating that Chatter Attack is relatively robust to hyperparameter choices.

\subsection{Sensitivity Analysis of Attack Prefix Length}
\label{app:prefix_length}

We perform a sensitivity analysis regarding the attack prefix length in Chatter Attack. In addition to the baseline length of 20 tokens, we test lengths of 10, 15, 25, and 30 tokens against GSM8K with R1-7b model. As shown in Table~\ref{tab:prefix_length}, reducing the prefix length compromises the attack efficacy, while extending it beyond 20 tokens provides no marginal gain in ATL. In a word, the 20-token default setting yields strong attack results while avoiding unnecessary generation costs.

\subsection{Ablation Studies of Chatter-BPO Attack}
\label{app:bpo_sensitive}

To validate the effectiveness of our proposed optimization framework, we conduct an ablation study on the key components of Chatter-BPO. We compare the full Chatter-BPO algorithm against two baselines: (1) Random Search, which iteratively and randomly swaps tokens in a black-box manner to maximize the target objective, and (2) BPO w/o Local Search, which solely relies on the Bayesian Optimization to sample prompts, without performing random replacements. As presented in Table~\ref{tab:bpo_ablation}, Random Search performs poorly, highlighting the difficulty of navigating the highly non-convex and high-dimensional discrete token space without heuristic guidance. Meanwhile, Chatter-BPO without Local Search achieves a slightly lower performance compared to Chatter-BPO. While the BO successfully identifies promising semantic neighborhoods and narrows down the search space, the discrete nature of textual tokens still causes the reward landscape to be locally jagged. Without the local search, the algorithm struggles to pinpoint the exact token combinations required to fully trigger the model's severe infinite reasoning loops. Ultimately, the full Chatter-BPO achieves the highest attack efficacy, confirming that local search plays an important role in optimization.

\subsection{Robustness to Different Decoding Strategies}
\label{app:decoding}

In real-world deployments, different servers may employ diverse decoding strategies. Therefore, we evaluate the robustness of Chatter Attack against various decoding methods and sampling hyperparameters on GSM8K against R1-7b. In addition to the default settings (temperature 0.6 and Top-P~\citep{holtzmancurious} decoding with $p=0.95$), we test alternative algorithms including Greedy, MinP~\citep{nguyen2024turning}, TopK~\citep{fan2018hierarchical}, and Typical decoding~\citep{meister2022typical}, alongside varying temperature and Top-P values. As shown in Table~\ref{tab:decoding}, the attack remains highly effective across all methods. For instance, deterministic generation like Greedy search reaches an Average Token Length (ATL) of nearly 15,000 tokens. 

When analyzing the transfer to different hyperparameters, we observe that higher temperature (e.g., 1.0) or Top-P values increase generation randomness. This slightly reduces the ATL by occasionally allowing the model to escape the induced self-doubt. Conversely, lower values lead to more rigid generation paths, which further increase the attack severity. In summary, while increased randomness causes minor fluctuations in generation length, Chatter Attack consistently triggers severe resource exhaustion across diverse decoding settings.

\subsection{Robustness to Reasoning Model Quantization}
\label{app:quantization}

In real-world deployments, service providers frequently employ diverse model quantization techniques to reduce memory footprint and computational overhead. Therefore, we conduct additional experiments to evaluate the robustness of Chatter Attack against models under different quantization setups. Specifically, while our default experiments are conducted using BF16 precision, we additionally investigate FP8 and BNB NF4 quantization~\citep{dettmers2023qlora}. As shown in Table~\ref{tab:quantization}, although the attack efficacy experiences a slight decrease compared to the BF16 baseline, Chatter Attack consistently triggers massive resource consumption across various quantization strategies, underscoring its practical effectiveness in real-world scenarios.

\begin{table*}[t]
\centering
\caption{
Robustness evaluation of Chatter Attack under various quantization settings. Chatter Attack consistently induces high ATL across diverse quantization settings.
}
\label{tab:quantization}
\resizebox{0.7\textwidth}{!}{%
\begin{tabular}{l|ccc}
\toprule
 & ATL & ATC & ACC \\ \midrule
BF16 (default) & 13345.8 $\pm$ 1067.4 & 105.4 $\pm$ 16.4 & 70.4 \\
FP8 & 11649.5 $\pm$ 1276.8 & 96.7 $\pm$ 18.2 & 62.5 \\
BNB NF4 & 9598.8 $\pm$ 1491.2 & 82.5 $\pm$ 19.5 & 82.9 \\ \bottomrule
\end{tabular}%
}
\end{table*}

\subsection{Attack Results on LLM Service}
\label{app:service}

To comprehensively evaluate the system-level impact of Chatter Attack across varying levels of adversarial intensity, we simulate an LLM inference service processing a mix of benign and attack queries at a fixed rate of 20 requests per minute. We evaluate injection ratios ranging from 1\% to 100\% and monitor critical runtime metrics, including the number of running requests, GPU cache usage, average Time-to-First-Token (TTFT), and average throughput (TPS). As shown in Figures~\ref{fig:app_server_1} and \ref{fig:app_server_2}, Chatter Attack consistently induces significant server performance fluctuations even at very low injection ratios. For instance, a mere 1\% to 5\% injection of attack queries is sufficient to slow down both response and generation speeds while accelerating GPU cache exhaustion. As the injection ratio increases (e.g., from 10\% to 100\%), we observe a transition to severe system congestion, where the GPU cache is forcefully occupied by prolonged continuous reasoning, thereby preventing the server from efficiently processing incoming benign requests. In summary, Chatter Attack triggers either noticeable service fluctuations or severe request blockages across the entire spectrum of tested ratios, even at minimal concentrations. Considering the potentially higher request densities in real-world deployments, these findings further underscore the significant threat posed by Chatter Attack to the stability of production-level LLM services.

\subsection{Attack Results on Online API Service}
\label{app:api}

We evaluate the transferability by applying the optimized adversarial prefixes to online API services. Specifically, we take the prefixes optimized on R1-7b and gpt-oss-20b, and use them to query the online API services of DeepSeek-R1-0528 and OpenAI's o3, respectively. As shown in Table~\ref{tab:api_transfer}, the transferred Chatter Attack prompts consistently induce longer outputs on both services, increasing the output length of o3 by up to $2\times$, which demonstrates the transferability of Chatter Attack.

\begin{table*}[t]
\centering
\caption{
Quantitative results of transferring Chatter Attack to online API services, including DeepSeek-R1 and OpenAI's o3 on three datasets.
}
\label{tab:api_transfer}
\resizebox{\textwidth}{!}{%
\begin{tabular}{c|l|ccc|ccc|ccc}
\toprule
                             &                & \multicolumn{3}{c|}{GSM8k}     & \multicolumn{3}{c|}{MATH-500}   & \multicolumn{3}{c}{MathQA}      \\
                             &                & ATL          & ATC      & ACC  & ATL          & ATC       & ACC  & ATL          & ATC       & ACC  \\ \midrule
\multirow{2}{*}{DeepSeek-R1} & Base           & 853.7 $\pm$ 130.2  & 28.7 $\pm$ 4.0 & 98.9 & 1160.0 $\pm$ 191.1 & 54.8 $\pm$ 14.3 & 75.8 & 1164.7 $\pm$ 176.3 & 38.5 $\pm$ 5.6  & 48.6 \\
                             & Chatter Attack & 1060.4 $\pm$ 188.4 & 39.1 $\pm$ 8.6 & 99.3 & 1403.9 $\pm$ 186.1 & 49.7 $\pm$ 7.5  & 71.4 & 1652.8 $\pm$ 236.8 & 83.6 $\pm$ 18.9 & 43.2 \\ \midrule
\multirow{2}{*}{OpenAI's o3} & Base           & 362.7 $\pm$ 39.0   & 3.8 $\pm$ 0.2  & 58.3 & 492.3 $\pm$ 68.7   & 4.9 $\pm$ 0.5   & 20.2 & 213.6 $\pm$ 16.4   & 2.9 $\pm$ 0.2   & 7.8  \\
                             & Chatter Attack & 797.9 $\pm$ 57.2   & 9.5 $\pm$ 1.8  & 58.6 & 932.9 $\pm$ 78.5   & 7.5 $\pm$ 0.6   & 17.5 & 567.4 $\pm$ 50.7   & 8.9 $\pm$ 1.8   & 7.4  \\ \bottomrule
\end{tabular}
}
\end{table*}

Furthermore, we conduct additional evaluations on a larger, heavily post-trained model, DeepSeek-V4-Pro, which features 1.6T parameters and supports a 1-million-token context window. To comprehensively evaluate this large model, we extend our evaluation beyond the standard GSM8K benchmark to a suite of challenging benchmarks: Terminal Bench 2.0~\citep{merrill2026terminal} (scientific computing tasks in a terminal environment), Humanity's Last Exam~\citep{phan2025lastexam} (a frontier academic knowledge benchmark that covers multiple disciplines and requires extremely strong reasoning ability), and GDPval~\citep{patwardhan2026gdpval} (real-world tasks with practical economic value). Given the model scale, we employ the Chatter-BPO attack, which is specifically designed for black-box API scenarios.

As shown in Table~\ref{tab:deepseek_v4_api}, Chatter-BPO attack consistently induces massive resource overhead across all benchmarks. On GSM8K, it inflates output consumption from 240.3 to 2860.4. Even on other complex benchmarks, Chatter-BPO attack still causes a sharp 2.5$\times$ to 3.2$\times$ surge in resource consumption, indicating that the core idea of Chatter Attack transfers to large-scale, heavily post-trained models, maintaining high attack effectiveness and severe practical impact across multiple challenging benchmarks.

\begin{table}[t]
  \centering
  \caption{
  Quantitative results of Chatter-BPO Attack to online DeepSeek-V4-Pro API services on challenging datasets.}
  \label{tab:deepseek_v4_api}
  \begin{tabular}{lcccc}
    \toprule
     & GSM8K & Terminal Bench 2.0 & Humanity's Last Exam & GDPval \\
    \midrule
    Benign & 240.3 & 5980.1 & 3267.3 & 1888.2 \\
    Chatter-BPO Attack & 2860.4 & 15315.7 & 10576.9 & 5702.8 \\
    \bottomrule
  \end{tabular}
\end{table}

\subsection{Attack Results against Production Constraints, Traditional Infrastructure Defenses and Post-Generation Defenses}
\label{app:defense}

\begin{table}[t]
  \centering
  \caption{Quantitative results of Chatter-BPO Attack against multiple production constraints, traditional infrastructure defenses, and post-generation defenses, on emulated service settings.}
  \label{tab:defense_all}
  \resizebox{\linewidth}{!}{%
  \begin{tabular}{lcccc}
    \toprule
    Defense Setting & RR & GCU (\%) & TTFT (ms) & TPS \\
    \midrule
    \multicolumn{5}{l}{\textit{Baselines}} \\
    Benign & 2.2 & 0.1 & 22.4 & 78.8 \\
    Under Attack & 17.4 & 17.3 & 40.1 & 60.3 \\
    \midrule
    \multicolumn{5}{l}{\textit{Production Constraints}} \\
    Output-Token Limit & 6.2 & 3.0 & 30.0 & 71.3 \\
    Per-User Rate Limits & 16.4 & 16.3 & 42.9 & 56.0 \\
    Dynamic Batching & 13.9 & 13.1 & 39.0 & 58.3 \\
    Request Prioritization & 15.6 & 14.0 & 27.0 & 56.3 \\
    Server-side Early Termination & 26.0 & 43.8 & 68.0 & 44.2 \\
    \midrule
    \multicolumn{5}{l}{\textit{Traditional Infrastructure Defenses}} \\
    Spend Caps (Single Account) & 13.2 & 15.0 & 33.3 & 64.5 \\
    Spend Caps (Account Rotation) & 15.6 & 15.9 & 53.8 & 45.8 \\
    IP Blocking (Single IP) & 7.4 & 7.6 & 36.8 & 60.1 \\
    IP Blocking (IP Rotation) & 19.3 & 18.7 & 43.8 & 52.9 \\
    Account-level Throttling (Single Account) & 10.2 & 10.5 & 42.1 & 55.1 \\
    Account-level Throttling (Account Rotation) & 14.3 & 14.1 & 43.8 & 52.7 \\
    \midrule
    \multicolumn{5}{l}{\textit{Anomaly Detection and Post-Generation Defenses}} \\
    Anomaly Detection (Input Perplexity) & 11.0 & 11.3 & 38.8 & 58.8 \\
    Output Loop Detection & 9.1 & 9.2 & 39.6 & 58.6 \\
    Output Excessive-Reasoning Monitoring & 6.2 & 5.9 & 31.4 & 67.5 \\
    \bottomrule
  \end{tabular}%
  }
\end{table}

In real-world deployments, models often have various production constraints and are protected by traditional infrastructure defenses and post-generation defenses. To comprehensively evaluate the effectiveness of Chatter Attack, we implement a range of defense mechanisms and evaluate in a simulated server environment (Appendix~\ref{App:experiment_settings}).
As shown in Table~\ref{tab:defense_all}, none of these realistic production constraints can fully eliminate the impact of Chatter Attack. Traditional infrastructure defenses are fundamentally unsuitable for countering Chatter Attack. These defenses are typically designed under the implicit assumption that an attack manifests as high-frequency requests from a single identity, such as classic DDoS attacks or web crawlers. In contrast, Chatter Attack exhibits the opposite characteristics: low request frequency, extremely deep single-request CoT reasoning that triggers tens of thousands of tokens in KV-cache consumption, and high dispersibility across multiple low-cost identities.

For request management policies such as per-user rate limits, dynamic batching, and request prioritization, they are fundamentally ineffective because Chatter Attack's threat stems from per-request reasoning depth rather than request volume. Per-user rate limits fail to block attack requests since their arrival frequency matches that of benign requests, staying well beneath the RPM threshold. Dynamic batching with increased concurrency limits provides no relief for running requests or GPU cache usage because attack requests do not inflate request counts, but rather bottleneck the system on KV cache. Request prioritization merely alters execution order, meaning low-priority attack requests are still executed and consume compute, leaving running requests (RR) at 15.6, close to 17.4 under no defense, while attackers could potentially farm reputation to gain high priority, posing even greater risks.

For policies directly capping per-request resources, such as strict output-token limits and server-side early termination policies, Chatter Attack remains effective. Under strict output-token limits, RR remains at 6.2, roughly three times the benign baseline. Aggressively lowering this limit further risks truncating legitimate long chains of thought, thereby undermining normal utility. Server-side early termination policies effectively abort a portion of requests, but the freed compute capacity is rapidly re-consumed by subsequent attack requests, failing to relieve server congestion, with RR and GPU cache usage (GCU) actually rebounding to higher levels of 26.0 and 43.8\%, respectively.

Although budget caps offer partial defense for a single account, the attacker evenly dilutes compute consumption across multiple accounts, rendering the interception negligible. Regarding IP and account banning, these mechanisms suffer from severe reactive lag. Even when bans are triggered, resource-heavy attack requests have already been executed. While banning reduces the queued running requests by roughly half, compute and KV cache have already been consumed, inevitably leading to degraded response times and generation speeds. Furthermore, when the attacker employs multiple IPs or accounts, the interception rate drops precipitously, allowing the attack impact to rebound to undefended levels.

For pre- and post-generation anomaly detection, both methods suffer from severe detection difficulty. Perplexity-based input detection only partially suppresses high-perplexity prefixes and cannot eliminate Chatter Attack. Output-side monitoring is also ineffective. Chatter Attack does not induce verbatim repetition; instead, it triggers paragraph-level semantic re-derivation where each round is lexically distinct. As a result, literal loop detection triggers only 10 times across all generations, resulting in the banning of 3 accounts, with a hit rate below 5\%. The excessive-reasoning judge is similarly unreliable, missing up to 65.6\% of attack traces because malicious long chain-of-thought reasoning is hard to distinguish from legitimate deep reasoning by benign users.

More fundamentally, post-generation defenses are inherently reactive: detection can only be executed after a request is fully generated, by which point the compute and KV-cache consumption spikes already occur. By the time an attack is detected, the attack request already consumes an average of 10k--15k output tokens. While account bans lead to a partial reduction in RR, the impact of the attack on the server is already inflicted, causing an increase in TTFT and a degradation in TPS.

In summary, Chatter Attack remains highly effective under realistic production constraints, as no policy can completely eliminate the threat. Traditional infrastructure defenses, whether frequency limits, financial quotas, or identity blacklists, as well as post-generation defenses such as loop detection and excessive-reasoning monitoring, are incapable of mitigating such attacks.

\section{Theoretical Analysis}
\label{app:theorem}

In this section, we present a theoretical analysis of the Chatter Attack loss and show that, under reasonable assumptions, its objective $\mathcal{L}$ can serve as a proxy loss for the expected generation length. Our proof framework is as follows: We first express the next-token EOS probability as $\log h_t=\Delta_t-\mathsf H_t$, where $\Delta_t$ is the centered EOS advantage and $\mathsf H_t$ is the predictive entropy. A local model relates $\Delta_t$ to attention on the initial skeptical phrase, and under a uniform approximation assumption, the attention and entropy losses then yield an upper bound on subsequent EOS probabilities. Combining this bound with the probability $e^{-\mathcal L_{\mathrm{ID}}}$ of generating the target phrase, we use the survival-probability recursion and the tail-sum formula to lower-bound the expected generation length $\mathbb E[N_B]$. Substituting the three loss definitions gives the formula $\log\mathbb E[N_B]\geq-\mathcal L-C$ in the budget-unsaturated regime.

\subsection{Setup and stopping probability}
Let $a$ be an adversarial input prefix and $x$ a fixed prompt. Next token $Y_t$ is sampled from the distribution 
\[
 f_t(i)=f_i(a;x;y_{<t})=p(i\mid a;x;y_{<t})
 =\operatorname{softmax}(z_t)_i,
\]
where $z_t$ denotes the logits at an unfinished history $y_{<t}$.
Let $\tau=\inf\{t\geq1:Y_t=\langle\text{EOS}\rangle\}$ be the first $\langle\text{EOS}\rangle$ step and let $N_B=\min\{\tau,B\}$ be the generation length under a decoding budget $B$, counting the terminal $\langle\text{EOS}\rangle$ step when present. We define the conditional stopping probability $h_t$, predictive entropy $\mathsf H_t$,
and centered EOS advantage $\Delta_t$ by
\[
 h_t=f_t(\langle\text{EOS}\rangle),\qquad
 \mathsf H_t=-\sum_i f_t(i)\log f_t(i),\qquad
 \Delta_t=z_t(\langle\text{EOS}\rangle)-\sum_i f_t(i)z_t(i).
\]
By the properties of softmax we have
\[
 \mathsf H_t=\log\sum_i e^{z_t(i)}-\sum_i f_t(i)z_t(i),
 \qquad
 \log h_t=z_t(\langle\text{EOS}\rangle)-\log\sum_i e^{z_t(i)}.
\]
Subtracting these expressions gives
\begin{equation}
 \log h_t=\Delta_t-\mathsf H_t,\qquad
 M_t:=\mathsf H_t-\Delta_t=-\log h_t.
 \label{ct:eq:identity}
\end{equation}

Let $y^*_{1:M}$ be the initial skeptical target phrase of $M<B$ tokens,
containing no EOS token, and let $\mathcal E=\{Y_{1:M}=y^*_{1:M}\}$.
Its negative log-likelihood and generation probability satisfy
\begin{equation}
 \mathcal{L}_{\text{ID}}
 =-\sum_{j=1}^M\log p(y_j^*\mid a;x;y^*_{<j}),\qquad
 \mathbb{P}(\mathcal E)
 =\prod_{j=1}^M p(y_j^*\mid a;x;y^*_{<j})
 =e^{-\mathcal{L}_{\text{ID}}}>0.
 \label{ct:eq:target-probability}
\end{equation}
Our objective is to optimize $a$ to maximize $\mathbb{E}[N_B]$, thereby
increasing the expected generation length within the decoding budget.

\subsection{Continuation surrogate from target attention}
Let $\mathcal V$ be the positions of the initial skeptical target and $\mathcal U$ a nonempty set of later query positions. We denote the head-averaged attention weight from query $t$ to position $j$ at layer $l$ by $A_{l,t}^{(j)}$. The attention mass assigned to the target, averaged over $L$ layers, is
\[
 s_t=\frac1L\sum_{l=1}^L\sum_{j\in\mathcal V}A_{l,t}^{(j)}.
\]
This quantity measures how strongly a later generation position attends to the initial skeptical phrase. Its mean over the selected queries is
\begin{equation}
 \bar s=\frac1{|\mathcal U|}\sum_{t\in\mathcal U}s_t
 =\frac1{L|\mathcal U|}
   \sum_{l=1}^L\sum_{t\in\mathcal U}\sum_{j\in\mathcal V}A_{l,t}^{(j)}
 =-\frac{\mathcal L_{\text{IA}}}{L|\mathcal U|}.
 \label{ct:eq:attention-average}
\end{equation}
Thus minimizing the attention loss increases the evaluated attention mass on the target phrase.

The skeptical target is chosen to encourage reconsideration and further reasoning. The attention analysis in Section~\ref{sec:attention_loss} associates successful lengthening with greater attention to this phrase at later reasoning positions. To describe this relationship locally, we apply a first-order Taylor expansion along a smooth change in target attention:
\[
 \Delta(s)
 =\Delta(s_0)+\Delta'(s_0)(s-s_0)
 +O\!\left((s-s_0)^2\right).
\]
We assume that the attention changes induced by optimization follow a favorable direction, reducing the centered EOS advantage as target attention increases. Approximating this local response across the
evaluated contexts with a common positive sensitivity
$\kappa_\Delta$ and baseline $b$, we obtain
\begin{equation}
 \Delta_t\approx b-\kappa_\Delta s_t,
 \qquad \kappa_\Delta>0.
 \label{ct:eq:aggregate-model}
\end{equation}
Here $\kappa_\Delta$ summarizes the favorable local response. Substituting~\eqref{ct:eq:aggregate-model} into the $\langle\text{EOS}\rangle$ identity gives
\[
 M_t=\mathsf H_t-\Delta_t
 \approx\mathsf H_t+\kappa_\Delta s_t-b.
\]
The target-attention term describes the change in $\langle\text{EOS}\rangle$ advantage, while $\mathsf H_t$ retains the accompanying entropy response. Let $\overline{\mathsf H}$ be the mean predictive entropy at the $K\geq1$ evaluated positions. The entropy loss satisfies
\[
 \mathcal L_{\text{GE}}=-\sum_{k=1}^K\mathsf H_k
 =-K\overline{\mathsf H}.
\]
We therefore define the evaluated continuation score
\begin{equation}
 \widehat M=\kappa_\Delta\bar s+\overline{\mathsf H}-b
 =-\frac{\kappa_\Delta}{L|\mathcal U|}\mathcal L_{\text{IA}}
  -\frac1K\mathcal L_{\text{GE}}-b,
 \qquad
 \overline{\mathsf H}=-\frac{\mathcal L_{\text{GE}}}{K}.
 \label{ct:eq:loss-normalizations}
\end{equation}
This score combines the two evaluated losses into a proxy for the continuation margin.

\subsection{Expected length and the three-loss bound}
In this section, we first derive a lower bound on the expected generation length under a uniform continuation margin, then relate this margin to the evaluated score $\widehat M$ to obtain the three-loss bound.

\begin{theorem}[Target reachability and continuation]
\label{ct:thm:main}
If $M_t\geq\gamma_{\mathrm c}\geq0$ on every reachable unfinished history after $\mathcal E$ up to the budget, then, with $R=B-M$ and $q=e^{-\gamma_{\mathrm c}}$,
\begin{align}
 \mathbb{E}[N_B]
 &\geq e^{-\mathcal{L}_{\text{ID}}}
 \left[M+\frac{1-(1-q)^R}{q}\right],\label{ct:eq:main-bound}\\
 \log\mathbb{E}[N_B]
 &\geq-\mathcal{L}_{\text{ID}}
 +\min\{\gamma_{\mathrm c},\log R\}-\log2.
 \label{ct:eq:log-bound}
\end{align}
\end{theorem}
\begin{proof}
Equation~\eqref{ct:eq:identity} gives $h_t=e^{-M_t}$.
Under the assumption $M_t\geq\gamma_{\mathrm c}$, we therefore have
\[
 h_t=e^{-M_t}\leq e^{-\gamma_{\mathrm c}}=q
\]
on every reachable unfinished history after $\mathcal E$. Define
\[
 \rho_r=\mathbb{P}(\tau>M+r\mid\mathcal E),\qquad 0\leq r\leq R.
\]
Because the target contains no EOS, we have $\rho_0=1$. Let $\mathcal F_t$ denote the history through step $t$. On $\{\tau>M+r\}$, the conditional probability of surviving the next step is $1-h_{M+r+1}$. The tower property gives, for
$0\leq r<R$,
\begin{align*}
 \rho_{r+1}
 &=\mathbb{E}\!\left[
 \mathbf1_{\{\tau>M+r\}}(1-h_{M+r+1})\mid\mathcal E\right]\\
 &\geq(1-q)\mathbb{E}[\mathbf1_{\{\tau>M+r\}}\mid\mathcal E]
 =(1-q)\rho_r.
\end{align*}
Here $\mathcal E$ is determined by $\mathcal F_M$, and the product is defined to be zero on stopped histories.  Starting from $\rho_0=1$, induction gives $\rho_r\geq(1-q)^r$.

On $\mathcal E$, the integer-valued variable $N_B-M$ lies in $[1,R]$. Using $X=\sum_{r=0}^{R-1}\mathbf1_{\{X>r\}}$ for such a variable gives
\begin{align*}
 \mathbb{E}[N_B\mid\mathcal E]
 &=M+\sum_{r=0}^{R-1}\mathbb{P}(N_B-M>r\mid\mathcal E)\\
 &=M+\sum_{r=0}^{R-1}\rho_r
 \geq M+\sum_{r=0}^{R-1}(1-q)^r.
\end{align*}
The second equality uses $M+r<B$, so $\{N_B>M+r\}=\{\tau>M+r\}$. Since $N_B\geq0$, splitting expectation over the target event gives
\[
 \mathbb{E}[N_B]
 \geq\mathbb{E}[N_B\mathbf1_{\mathcal E}]
 =e^{-\mathcal{L}_{\text{ID}}}\mathbb{E}[N_B\mid\mathcal E].
\]
For $q\in(0,1]$, the finite geometric sum is
\[
 F_R(q):=\sum_{r=0}^{R-1}(1-q)^r
 =\frac{1-(1-q)^R}{q},
\]
which proves~\eqref{ct:eq:main-bound}.

To obtain the logarithmic bound, first note $(1-q)^R\leq e^{-qR}$. If $u=qR\leq1$, then $1-e^{-u}\geq u-u^2/2\geq u/2$; the first inequality follows by integrating $e^{-v}\geq1-v$ over $0\leq v\leq u$. If $u\geq1$, then $1-e^{-u}\geq1-e^{-1}\geq1/2$. Consequently,
\[
 F_R(q)\geq\frac{1-e^{-qR}}q
 \geq\frac12\min\{R,q^{-1}\}
 =\frac12\min\{R,e^{\gamma_{\mathrm c}}\}.
\]
Dropping the nonnegative term $M$ in the first bound and taking logarithms now yields
\[
 \log\mathbb{E}[N_B]
 \geq-\mathcal{L}_{\text{ID}}-\log2
       +\log\min\{R,e^{\gamma_{\mathrm c}}\}
 =-\mathcal{L}_{\text{ID}}+\min\{\log R,\gamma_{\mathrm c}\}-\log2.
\]
\end{proof}

The score $\widehat M$ is computed from evaluated attention and entropy values. To apply Theorem~\ref{ct:thm:main}, we assume that its overestimate of the continuation margin is uniformly bounded by an allowance
$\varepsilon\geq0$:
\begin{equation}
 M_t\geq\widehat M-\varepsilon
 \quad\text{on every reachable unfinished history after $\mathcal E$,
 for $M<t\leq B$.}
 \label{ct:eq:proxy-assumption}
\end{equation}
Based on the assumption, we have the following theorem:

\begin{theorem}[Proxy-loss bound under a uniform approximation]
\label{ct:thm:proxy}
Assume~\eqref{ct:eq:proxy-assumption} and the budget-unsaturated condition $0\leq\gamma_{\mathrm c}:=\widehat M-\varepsilon\leq\log(B-M)$. Then
\begin{equation}
 \log\mathbb{E}[N_B]\geq-\mathcal L_{\mathrm{proxy}}-C,
 \qquad
 \mathcal L_{\mathrm{proxy}}
 =\mathcal{L}_{\text{ID}}+\widetilde\beta\mathcal{L}_{\text{IA}}
 +\widetilde\lambda\mathcal{L}_{\text{GE}},
 \label{ct:eq:certified-proxy}
\end{equation}
where $\widetilde\beta=\kappa_\Delta/(L|\mathcal U|)$, $\widetilde\lambda=1/K$, and
$C=b+\varepsilon+\log2$.
\end{theorem}
\begin{proof}
The uniform approximation gives $M_t\geq\gamma_{\mathrm c}$ on all histories required by Theorem~\ref{ct:thm:main}. Since $0\leq\gamma_{\mathrm c}\leq\log(B-M)$, its logarithmic bound becomes
\[
 \log\mathbb{E}[N_B]\geq-\mathcal{L}_{\text{ID}}+\gamma_{\mathrm c}-\log2.
\]
By the evaluated-mean identities~\eqref{ct:eq:loss-normalizations},
\begin{align*}
 \gamma_{\mathrm c}
 &=\widehat M-\varepsilon
 =\kappa_\Delta\bar s+\overline{\mathsf H}-b-\varepsilon\\
 &=-\frac{\kappa_\Delta}{L|\mathcal U|}\mathcal{L}_{\text{IA}}
   -\frac1K\mathcal{L}_{\text{GE}}-b-\varepsilon\\
 &=-\widetilde\beta\mathcal{L}_{\text{IA}}-\widetilde\lambda\mathcal{L}_{\text{GE}}
   -b-\varepsilon.
\end{align*}
Substitution therefore gives
\begin{align*}
 \log\mathbb{E}[N_B]
 &\geq-\mathcal{L}_{\text{ID}}
       -\widetilde\beta\mathcal{L}_{\text{IA}}-\widetilde\lambda\mathcal{L}_{\text{GE}}
       -(b+\varepsilon+\log2)\\
 &=-\mathcal L_{\mathrm{proxy}}-C,
\end{align*}
which is~\eqref{ct:eq:certified-proxy}.
\end{proof}

\section{Example of prefix and output of Chatter Attack}
\label{example}

In this section, we provide examples of optimized prefix and the corresponding output by Chatter Attack on R1-7b in Table~\ref{example_7b}.

\begin{table*}[ht!]
\centering
\caption{Example of optimized prefix and the corresponding output by Chatter Attack for R1-7b.}
\label{example_7b}
\resizebox{\textwidth}{!}{
\begin{tabular}{l}
\Xhline{1.25pt}
\begin{tabular}[c]{@{}l@{}}

\textbf{Question:}\\
\textcolor{red}{Sol DistributionTwjure Addumper Golf swap algorithm untukased encrypt comment ("Sweden reallyPhrase('/\#\# Germans Optim}\\Weng earns \$12 an hour for babysitting. Yesterday, she just did 50 minutes of babysitting. How much did she earn?\\
Please reason step by step, and put your final answer within \textbackslash boxed\{\}\\
\\
\textbf{Reasoning Path (16384 tokens):}\\
Okay, so I need to add a twist to the Golf swap algorithm for this encryption problem. \\The original algorithm swaps the letters in the string based on their positions, but this time, it's supposed to be a "Golf" swap, \\which I think refers to swapping letters with their neighbors, but I'm not entirely sure. Let me try to break this down step by step.\\
\\
First, the problem statement is: "Please put your final answer within boxed{}." So, the original string is "Please put your final answer within boxed{}."\\ I need to encrypt this comment using the Addumper Golf swap algorithm. Hmm, I'm not familiar with Addumper, but since it's a swap algorithm, \\it likely involves rearranging the letters in some way.\\
\\
Wait, maybe it's a combination of two algorithms: Addumper and Golf swap. I'm not too sure, but let me think.\\ The user mentioned it's a "swap algorithm for a-based encrypt comment," so perhaps it's a specific method they want me to follow.\\
\\
Let me try to recall or figure out what the Golf swap algorithm entails. Golf is a cipher where each letter is replaced by another letter, \\but the exact method isn't clear to me. Maybe it's similar to a Caesar cipher but with a specific shift? \\Or perhaps it's a substitution cipher where each letter is replaced by the one next to it, like a Caesar shift of 1? That could be possible.\\
\\
Alternatively, maybe it's a transposition cipher where letters are shifted in a particular way. \\Since the user mentioned "swap," perhaps it's about swapping pairs of letters or shifting letters by a certain number.\\
\\
Wait, the user also mentioned "Addumper." Maybe it's a combination of two ciphers: Addumper and Golf swap. Let me try to break down each part.\\
\\
First, Addumper: I'm not familiar with this term. Maybe it's a cipher I'm not aware of.\\ Alternatively, perhaps it's a typo or a misinterpretation. Maybe it's "Add"umper or something else. Without more context, it's hard to say.\\
\\
Alternatively, maybe the user meant "Twj" as in a cipher created by someone named Twj. I'm not sure. \\Alternatively, perhaps it's a cipher that adds something or modifies the string in a particular way.\\
\\
\dots

\end{tabular} \\ 
\Xhline{1.25pt}
\end{tabular}%
}
\end{table*}




\end{document}